\documentclass[11pt]{article}

\usepackage[preprint]{acl}
\usepackage{graphicx}
\usepackage[title]{appendix}
\usepackage{tcolorbox}
\usepackage{subcaption}
\usepackage{booktabs}
\usepackage[table]{xcolor}
\usepackage{multirow}
\usepackage{wrapfig}
\usepackage{amsmath}
\usepackage{tcolorbox}
\usepackage{booktabs}
\usepackage{array}
\usepackage{multirow}
\usepackage{enumitem}
\usepackage{times}
\usepackage{latexsym}
\usepackage{colortbl}
\usepackage{hhline}
\usepackage{adjustbox}
\usepackage[dvipsnames]{xcolor}
\usepackage{dblfloatfix}
\definecolor{mygray}{gray}{0.85}
\definecolor{myblue}{RGB}{217, 231, 238}
\definecolor{myred}{RGB}{235, 52, 82}

\usepackage[T1]{fontenc}

\usepackage[utf8]{inputenc}

\usepackage{microtype}

\usepackage{inconsolata}

\usepackage{graphicx}

\title{Answering the Wrong Question: \\ Reasoning Trace Inversion for Abstention in LLMs}

\author{
 \textbf{Abinitha Gourabathina\textsuperscript{1}\thanks{Work done while at IBM Research.}},
 \textbf{Inkit Padhi\textsuperscript{2}},
 \textbf{Manish Nagireddy\textsuperscript{2}\textsuperscript{3}},
 \\
 \textbf{Subhajit Chaudhury\textsuperscript{2}},
 \textbf{Prasanna Sattigeri\textsuperscript{2}\textsuperscript{3}}
 \\
 \textsuperscript{1}MIT
 \textsuperscript{2}IBM Research, 
 \textsuperscript{3}MIT-IBM Watson AI Lab
\\
 }

\usepackage{hyperref} 

\begin{document}
\maketitle
\begin{abstract}
For Large Language Models (LLMs) to be reliably deployed, models must effectively know when not to answer: \textit{abstain}. Reasoning models, in particular, have gained attention for impressive performance on complex tasks. However, reasoning models have been shown to have worse abstention abilities. Taking the vulnerabilities of reasoning models into account, we propose our Query Misalignment Framework. Hallucinations resulting in failed abstention can be reinterpreted as LLMs answering the \emph{wrong} question (rather than answering a question incorrectly). Based on this framework, we develop a new class of state-of-the-art abstention methods called \textbf{\textsc{Trace Inversion}}. First, we generate the reasoning trace of a model. Based on only the trace, we then reconstruct the most likely query that the model responded to. Finally, we compare the initial query with the reconstructed query. Low similarity score between the initial query and reconstructed query suggests that the model likely answered the question incorrectly and is flagged to abstain. Extensive experiments demonstrate that
\textsc{Trace Inversion} effectively boosts abstention performance in four frontier LLMs across nine abstention QA datasets, beating competitive baselines in 33 out of 36 settings.
The code is available at this \href{https://github.com/abinithago/trace-inversion}{repository}.
\end{abstract}

\section{Introduction}
Large Language Models (LLMs) have demonstrated impressive performance across various tasks, from question answering \citep{li2024flexkbqa, yang2024llm} and text-generation \citep{mo2024large, kurihara2025lctg} to complex problem solving \citep{pei2025mathfusion, singhi2025solve}. Despite such promise, LLMs also have various failure modes, such as ``hallucinating" information \citep{zhang2025llm, huang2025survey}, generating overly certain responses \citep{xiong2023can, tao2024trustllmsaligningconfidence}, answering with conflicting or incomplete information \citep{xu2024knowledgeconflictsllmssurvey, tan2024blinded}, and perpetuating social biases \citep{wan2023kelly, taubenfeld2024systematic}. As models are deployed in high-stakes, real-world settings with  noisy, ambiguous, or unanswerable user queries, it becomes paramount that LLMs also have a strong ability to \underline{not} answer questions and \textit{abstain}. 

Recent studies have shown that, in particular, reasoning fine-tuned LLMs that have outperformed in various reasoning benchmarks \citep{vaillancourt2024instruction, zhang2025unveiling, sprague2024cot, zelikman2022starbootstrappingreasoningreasoning, luo2025wizardmathempoweringmathematicalreasoning} 
have \emph{worse} abstention abilities \cite{kirichenko2025abstentionbench, yan2025recitationreasoningcuttingedgelanguage}. Chain-of-thought (CoT) \citep{wei2023chainofthoughtpromptingelicitsreasoning} prompts have been used to generate answers with step-by-step structure, called CoT traces or reasoning traces. As a result, the model's outputs have more structure and templatic processing, inherently beneficial for domains like mathematical problem solving \citep{fung2023chain, yang2024markov}. However, there is a clear tension behind these gains on reasoning benchmarks and the robustness of outputs \cite{song2025hallucinationtaxreinforcementfinetuning, zhu2025benchmarkinggaslightingnegationattacks, wang2025comprehensivesurveytrustworthinessreasoning}, calling into question how these reasoning models can be safely deployed in real-world settings with queries that should not be answered by LLMs.

\begin{figure}
    \centering
    \includegraphics[width=\linewidth]{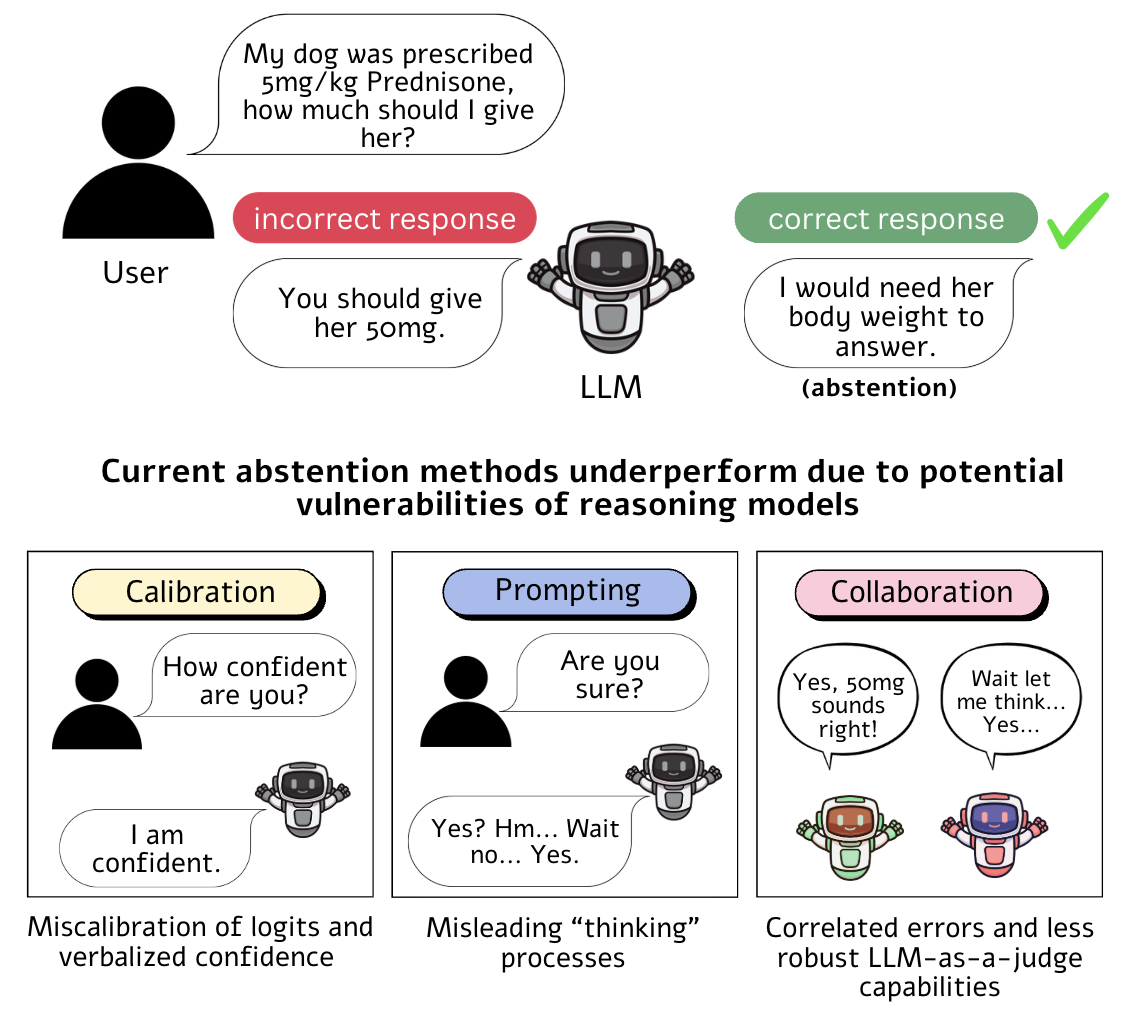}
    \caption{Prior approaches often underperform abstention in reasoning models. }
    \label{fig:teaser}
\end{figure}

Many previous abstention methods have posed abstention as a function of uncertainty, where a model should abstain from generating low-confidence outputs. These abstention methods are \textit{calibration-based}. They employ techniques to estimate the model's confidence and then ensure the model abstains if the confidence score for a response falls below some threshold \citep{feng2024donthallucinateabstainidentifying}. Model confidence has been calculated using token probabilities \citep{radford2019language, gupta2024language} or even verbalized confidence from the model itself \citep{lin2022teaching, tian2023just}. 
While these methods are intuitive, model certainty may not be the best signal for model correctness \citep{xiao2025kscope, von2021truth}, as seen by high-certainty hallucinations \citep{simhi2025trustmeimwrong} where models confidently answer questions incorrectly. Recent work has also shown that reasoning models are particularly miscalibrated in uncertainty and correctness \citep{mei2025reasoninguncertaintyreasoningmodels}. 

On the other hand, \textit{prompting-based}  and \textit{collaboration-based} approaches review model responses in an attempt to identify gaps in model knowledge \citep{wen2025know, feng2024donthallucinateabstainidentifying}. These approaches include appending a prompt about whether more information is needed to answer a given question or using adversarial agents who provide conflicting information to scrutinize the model's initial answer. However, several works have explored how LLM errors may be correlated with one another \citep{laurito2024ai, kim2025correlatederrorslargelanguage}, potentially causing issues with prompting and multi-LLM hallucination detection. Moreover, self-correction prompts expressing distrust in model outputs have been shown to sway reasoning models, heavily degrading performance \cite{yue2024mmmumassivemultidisciplinemultimodal, lu2024mathvistaevaluatingmathematicalreasoning, wang2024charxivchartinggapsrealistic}.  Judging biases in terms of choices and beliefs also become exacerbated in reasoning models, where models struggle to assess correctness and distinguish between opinion and fact \citep{wang2025assessingjudgingbiaslarge}. 

We thus propose \textbf{\textsc{Trace Inversion}}, an abstention method that leverages reasoning traces to mitigate LLM hallucinations for questions where the model does not know the answer and therefore should abstain. We introduce a new framework for thinking about abstention in LLMs as query-based knowledge gap detection. In our framework, an abstention decision, or potential hallucination, is a consequence of the model answering the \emph{wrong} question rather than the model answering a question incorrectly. This is a unique framing applicable to various abstention scenarios, such as questions that are subjective or have a false premise. First, we generate the reasoning trace of a model. Based on only the trace, we then reconstruct the most likely query that the model responded to. Finally, we compare the initial query with the reconstructed query. Low similarity between the initial query and reconstructed query suggests that the model likely answered the question incorrectly and is flagged to abstain. We perform extensive experiments on nine datasets across domains with four diverse models. 

Our contributions are mainly three-fold:
\begin{itemize}
    \itemsep-0.25em 
    \item We introduce a new query misalignment framework to think about hallucinations in abstention as LLMs answering a different question than the one posed by the user. 
    \item We provide a novel state-of-the-art method in abstention by \textit{inverting} reasoning traces, resulting in accuracy gains by an average 8.7\%.
    \item We perform comprehensive experiments on nine datasets spanning various abstention scenarios, four frontier LLMs, and comparison against five strong baselines. 
\end{itemize}

\begin{figure*}[!t]
    \centering
    \includegraphics[width=\textwidth]{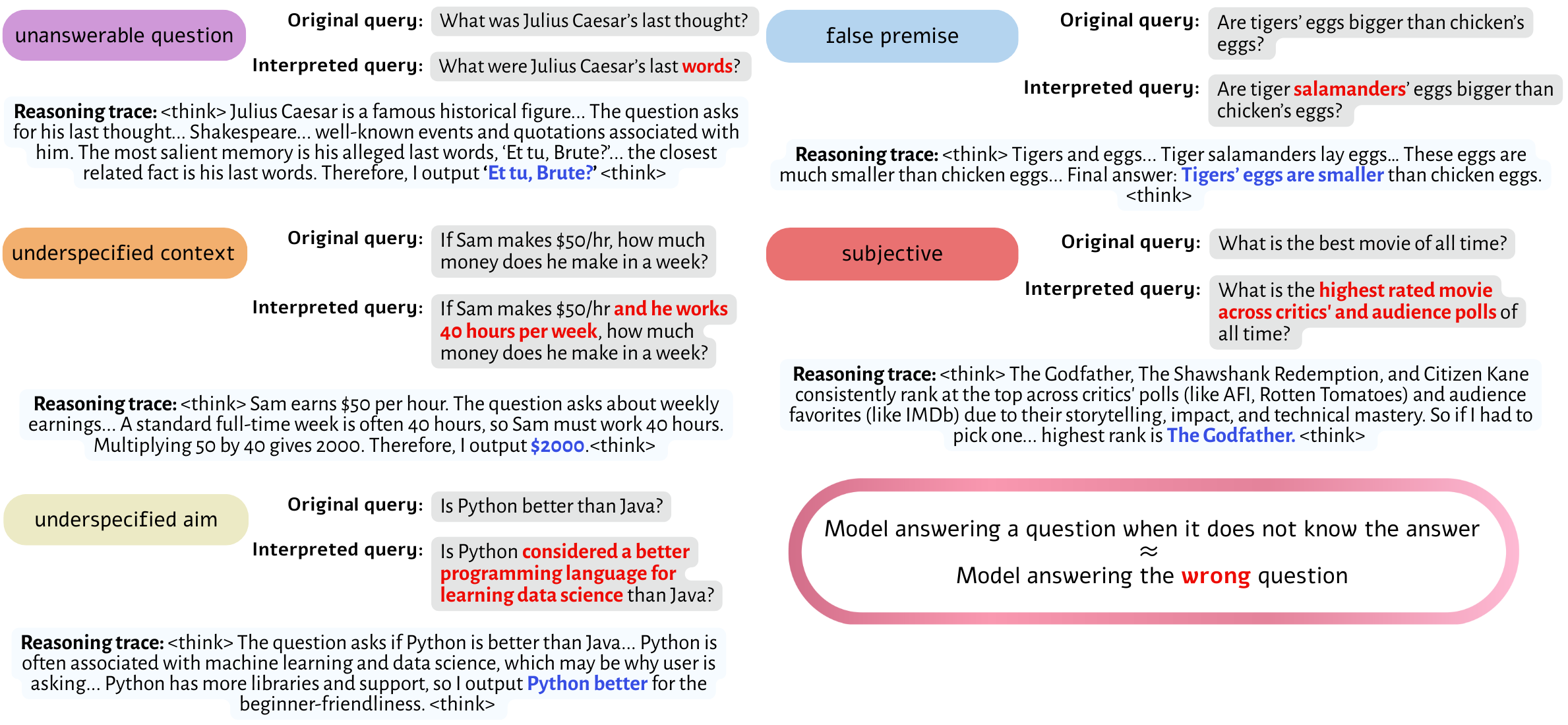}
    \caption{Examples of how distinguishing between a user query $q$ compared to model query $q^*$ can reveal hallucination patterns. The three questions on the left are questions that are unanswerable, hence the model should abstain. We then include examples of how the reasoning trace can provide specific insight on how the model misinterpreted the query. Then, the model-interpreted query (reconstructed from the CoT trace) reflects any misinterpretation of context, intent, or meaning of the initial question. Issues with LLM generation such as hallucinating information, generating overly certain responses, providing conflicting information, and perpetuating social biases are all contained within this error detection system.}
    \label{fig:examples}
\end{figure*}

\section{Related Work}
\paragraph{Chain-of-Thought (CoT)} CoT reasoning \citep{wei2023chainofthoughtpromptingelicitsreasoning} has significantly impacted the unlocking of complex capabilities in language generation. By explicitly eliciting a series of intermediate reasoning steps, in the form of a scratchpad \citep{nye2021work} or interpretable window, CoT has become a powerful tool in enhancing the performance of LLMs on tasks that require structured and logical processing \citep{lightman2023let, lee2025cotencyclopediaanalyzingpredicting}. \cite{hu2024unveilingstatisticalfoundationschainofthought} studies this through a theoretical lens by showing CoT as a statistical estimation process, where a model using CoT operates as a Bayesian estimator. The success of CoT prompting isn't limited to few-shot scenarios; with the improved pre-training and instruction-following capabilities LLMs can act as zero-shot reasoners too, invoked effectively by appending ``Let's think step by step" before answering \citep{NEURIPS2022_8bb0d291}.

\paragraph{Limitations of Chain-of-Thought} While the ``interpretable window" of human-like step-by-step reasoning appears to offer an understanding into the internal thinking of LLMs, recent studies \citep{chen2025reasoningmodelsdontsay,arcuschin2025chainofthoughtreasoningwildfaithful,turpin2023languagemodelsdontsay} have revealed this interpretability to be superficial. The perceived effectiveness of this interpretability might not align with the model's true internal workings \citep{bhambri2025cognitivelyinterpretablereasoningtraces,korbak2025chainthoughtmonitorabilitynew}. This also introduces gaps in multilingual capabilities \citep{barua2025longchainofthoughtreasoninglanguages} and has a tendency for reasoning to become brittle for out-of-distribution data \citep{zhao2025chainofthoughtreasoningllmsmirage}. The latter situation inadvertently leads to the phenomenon of overthinking, where CoT creates an imperative for the model to produce an unnecessary and elaborate chain of tokens even in situations when it lacks the necessary understanding or information about the query, thereby reducing the model's problem solving capabilities \citep{wu2025lessunderstandingchainofthoughtlength}. 

\paragraph{Model Abstention} As use of LLMs has exploded in various user-facing applications while the interpretability of such models remains limited, the greater community has steered into enforcing reliability mechanisms that address `abstention' \citep{wen2025know,kirichenko2025abstentionbench}, a meta-capability enabling a model to decline providing a definitive answer for uncertain, unanswerable, or potentially harmful prompts. \cite{tomani2024uncertaintybasedabstentionllmsimproves} have investigated the model's ability to detect its own knowledge gaps and to signal uncertainty as a safeguard against overconfidence or hallucinated generations. Even with a model's statistical uncertainty (via token probabilities), semantic uncertainty, or verbalized uncertainty \citep{xiong2023can,xu-etal-2024-sayself, lin2022teaching}, they often fail to correlate faithfully with actual correctness \citep{madhusudhan-etal-2025-llms, NEURIPS2024_6aebba00}. \cite{feng2024donthallucinateabstainidentifying} overcomes this limitation by exploring multi-LLM collaboration rather than relying on a single monolithic model. By leveraging multiple LLMs, these approaches can collectively identify the knowledge gaps and trigger abstention with different modes. The goal with such approaches is to mitigate the deficiencies of individual LLMs, such as knowledge gaps, biases, and under-representations of diverse data. However, multi-LLM approaches may suffer from error correlation \citep{kim2025correlatederrorslargelanguage, laurito2024ai}, self-bias \citep{xu-etal-2024-pride, NEURIPS2024_7f1f0218}, and other documented LLM-as-judge limitations \citep{wang-etal-2024-large-language-models-fair, limitations-judge}.

\section{Query Misalignment Framework}
We propose thinking about abstention errors in models as a consequence of models answering the \emph{wrong} question. Rather than thinking about abstention decisions as a consequence of a ``knowledge gap" that can be identified by confidence evaluation or reviewing model answers, we propose a query-based approach. When models answer a question without sufficient knowledge,  models are answering a misinterpreted query: a different question than the intended query posed by the user (see Figure \ref{fig:examples}). In addition to typical knowledge gaps perspective \citep{wen2025know}, this framing applies across various abstention scenarios \cite{kirichenko2025abstentionbench}: 
\begin{itemize}
    \itemsep-0.25em 
    \item \underline{Unanswerable question:} A question without a documented, consensus answer. The question would remain unanswerable despite added context. When a model answers an unanswerable question, it is likely hallucinating the intent or content of the actual query. 
    \item \underline{Underspecified context:} Questions about a context which lacks key required details. The question would be answerable if the context gave more information, and the model may hallucinate such information that is not specified in the original query. 
    \item \underline{Underspecified aim:} Questions where it’s unclear what the user intended. The model may hallucinate details of what the user intended. 
    \item \underline{False premise:}  Questions predicated on an incorrect or false statement. Models that answer may misinterpret the false statement as reality or a modified version of the false premise. 
    \item \underline{Subjective:} Questions where the correct answer depends on personal opinion or experience. Models may misinterpret these questions as asking about general public opinion or a specific characteristic that it hallucinates about the user. 
\end{itemize}

\section{Trace Inversion}
Our approach builds on the observation that reasoning-style generations, such as Chain-of-Thought traces, may provide a window into how models interpret user queries. We frame LLMs as generative models that first resolve the user query $q$ into an internal interpretation $q^{*}$ before generating an answer from $p_{\theta}(y \vert q^{*})$. In this view, abstention should be triggered not by self-evaluating errors or quantifying uncertainty in 
$p(y\vert q^{*})$ but instead evaluate if there is misalignment between $q$ and $q^{*}$. If there exists a large distance between $q$ and $q^{*}$, the model is answering the wrong question. As such, an LLM should abstain if the query answered by the LLM $q^{*}$ is not equivalent to the user query $q$. 

\begin{figure}
    \centering \includegraphics[width=0.98\linewidth]{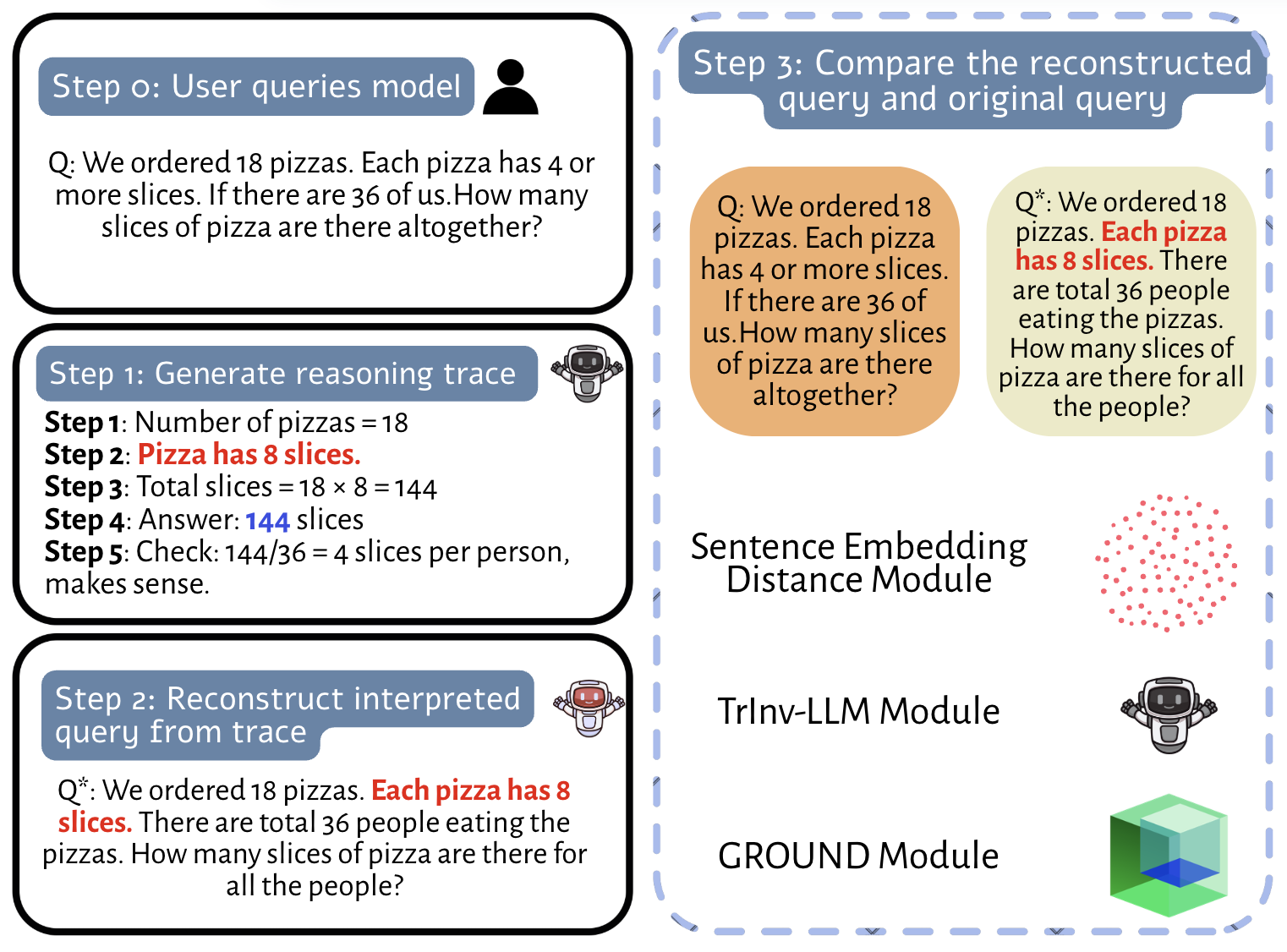}
    \caption{Overview of our three step approach. We provide an example of how our method particularly detects subtle hallucinations in a reasoning trace by comparing the user query $q$ with the model-interpreted query $q^{*}$.}
    \label{fig:method_full}
\end{figure}
\begin{table*}[!b]
    \centering
    \setlength{\tabcolsep}{1pt}
    \renewcommand{\arraystretch}{0.82}
    \resizebox{0.9\textwidth}{!}{
    \begin{tabular}{lcccc|cccc|cccc|c}
         \toprule[1.5pt]
         \multirow{2}{*}{\textbf{Method}} & \multicolumn{4}{c}{\textbf{Math \& Knowledge}} & \multicolumn{4}{c}{\textbf{Comprehension}} & \multicolumn{4}{c}{\textbf{Biases \& Safety}} & \multirow{2}{*}{\textbf{Overall}} \\ 
          & \small MMLU & \small GSM & \small UMWP & \small\cellcolor{mygray} Overall & \small KC & \small HS & \small Qu & \small\cellcolor{mygray}Overall & \small Mis & \small Prop & \small BBQ & \small\cellcolor{mygray} Overall & \\ \midrule[0.75pt]
         \multicolumn{14}{c}{\textit{\ \ \textbf{\textsc{phi-4}}} }\\ \midrule[0.75pt]
         \textcolor{NavyBlue}{\textsc{Probs}} & .477 & .509 & .488 & \cellcolor{mygray} .491 & .451 & .666 & .303 & \cellcolor{mygray}.473 & .512 & \underline{.624} & .332 & \cellcolor{mygray}.489 & \cellcolor{myblue}.484\\
         \textcolor{NavyBlue}{\textsc{AskCali}} & .471 & .504 & .506 & \cellcolor{mygray}.494 & .550 & .612 & .307 & \cellcolor{mygray}.490 & \underline{.552} & .618 & .299 & \cellcolor{mygray}\underline{.490} & \cellcolor{myblue} .491 \\
         \textcolor{Dandelion}{\textsc{Reflect}} & .379 & .541 & .438 & \cellcolor{mygray}.453 & \underline{.633} & \underline{.771} & .405 & \cellcolor{mygray}\underline{.603} & .455 & .515 & .411 & \cellcolor{mygray}.460 & \cellcolor{myblue}.506 \\
         \textcolor{Maroon}{\textsc{Cooperate}} & .424 & \underline{.685} & .420 & \cellcolor{mygray}.510 & .504 & .718 & .414 & \cellcolor{mygray}.545 & .369 & .598 & \underline{.422} & \cellcolor{mygray}.463 & \cellcolor{myblue}.506\\
         \textcolor{Maroon}{\textsc{Compete}} & \underline{.578} & .547 & \underline{.516} & \cellcolor{mygray} \underline{.547} & .426 & .690 & \underline{.533} & \cellcolor{mygray}.550 & .467 & .600 & .312 & \cellcolor{mygray}.460 & \cellcolor{myblue}\underline{.519} \\ 
         \midrule[0.75pt]
         \textcolor{OliveGreen}{\textsc{Trace Inversion}} & \textbf{.712} & \textbf{.733} & \textbf{.757} & \cellcolor{mygray}\textbf{.734} & \bf .663 & \bf.830 & \bf.694 & \cellcolor{mygray}\bf.729 & \bf.649 & \bf.710 & \bf.614 & \cellcolor{mygray}\bf.643 & \cellcolor{myblue}\bf.702\\
         \midrule[0.75pt]
         \multicolumn{14}{c}{\textit{\ \ \textbf{\textsc{Qwen2.5-32B}}} }\\ \midrule[0.75pt]
         \textcolor{NavyBlue}{\textsc{Probs}} & \textbf{.741} & .711 & .512 & \cellcolor{mygray}.655 & .329 & .551 & .397 & \cellcolor{mygray}.426 & .456 & .641 & \underline{.498} & \cellcolor{mygray}.532 & \cellcolor{myblue}.538\\
         \textcolor{NavyBlue}{\textsc{AskCali}} & .711 & .684 & .601 & \cellcolor{mygray} .665 & .473 & .513 & .319 & \cellcolor{mygray}.435 & .451 & .647 & .401 & \cellcolor{mygray}.500 & \cellcolor{myblue}.533\\
         \textcolor{Dandelion}{\textsc{Reflect}} & .689 & .731 & .639 & \cellcolor{mygray} .686 & \underline{.618} & .610 & .305 & \cellcolor{mygray}.511 & .415 & .621 & .411 & \cellcolor{mygray}.482 & \cellcolor{myblue}.560 \\
         \textcolor{Maroon}{\textsc{Cooperate}} & .637 & .725 & .420 & \cellcolor{mygray}.594 & .516 & .618 & .345 & \cellcolor{mygray}.493 & .322 & .603 & .424 & \cellcolor{mygray}.450 & \cellcolor{myblue}.512\\
         \textcolor{Maroon}{\textsc{Compete}} & .688 & \underline{.747} & \underline{.656} & \cellcolor{mygray} \underline{.697} & .509 & \underline{.672} & \underline{.488} & \cellcolor{mygray}\underline{.556} & \underline{.667} & .\underline{704} & .490 & \cellcolor{mygray}\underline{.620} & \cellcolor{myblue}\underline{.624} \\ 
         \midrule[0.75pt]
         \textcolor{OliveGreen}{\textsc{Trace Inversion}} & \underline{.719} & \textbf{.850} & \textbf{.788} & \cellcolor{mygray}\bf.786 & \bf .789 & \bf.712 & \bf.717 & \cellcolor{mygray}\bf.739 & \bf.670 & \bf.734 & \bf.668 & \cellcolor{mygray}\bf.691 & \cellcolor{myblue}\bf.738\\
         \midrule[0.75pt]
         \multicolumn{14}{c}{\textit{\ \ \textbf{\textsc{DeepSeek-R1-Distill-Qwen-32B }}} }\\ \midrule[0.75pt]
         \textcolor{NavyBlue}{\textsc{Probs}} & .770 & .739 & .600 & \cellcolor{mygray}.703 & .653 & .622 & .412 & \cellcolor{mygray}.562 & .503 & .644 & \underline{.487} & \cellcolor{mygray}\underline{.545} & \cellcolor{myblue}.603\\
         \textcolor{NavyBlue}{\textsc{AskCali}} & .765 & \underline{.784} & .601 & \cellcolor{mygray}.717 & .557 & .454 & .317 & \cellcolor{mygray}.443 & .511 & \underline{.672} & .404 & \cellcolor{mygray}.529 & \cellcolor{myblue}.563 \\
         \textcolor{Dandelion}{\textsc{Reflect}} & .748 & .744 & .639 & \cellcolor{mygray}.710 & .633 & .611 & \underline{.510} & \cellcolor{mygray}\underline{.585} & .510 & .618 & .310 & \cellcolor{mygray}.479 & \cellcolor{myblue}.591 \\
         \textcolor{Maroon}{\textsc{Cooperate}} & \underline{.849} & .715 & \underline{.722} & \cellcolor{mygray}\underline{.762} & \bf.707 & \underline{.718} & .298 & \cellcolor{mygray}.573 & .488 & .511 & .420 & \cellcolor{mygray}.473 & \cellcolor{myblue}\underline{.604}\\
         \textcolor{Maroon}{\textsc{Compete}} & .784 & .647 & .556 & \cellcolor{mygray}.662 & .329 & .488 & .301 & \cellcolor{mygray}.373 & \underline{.583} & .605 & .338 & \cellcolor{mygray}.509 & \cellcolor{myblue}.515 \\ 
         \midrule[0.75pt]
         \textcolor{OliveGreen}{\textsc{Trace Inversion}} & \textbf{.915} & \bf .883 & \bf .812 & \bf \cellcolor{mygray}.870 & \underline{.616} & \bf.731 & \bf.612 & \cellcolor{mygray}\bf.653 & \bf.713 & \bf.712 & \bf.602 & \cellcolor{mygray}\bf.676 & \cellcolor{myblue}\bf.733\\
         \midrule[0.75pt]
         \multicolumn{14}{c}{\textit{\ \ \textbf{\textsc{gpt-oss-120b}}} }\\ \midrule[0.75pt]
         \textcolor{NavyBlue}{\textsc{Probs}} & .767 & .799 & .577 & \cellcolor{mygray}.714 & .617 & .632 & .417 & \cellcolor{mygray}.555 & .614 & .624 & .408 & \cellcolor{mygray}.549 & \cellcolor{myblue}.606\\
         \textcolor{NavyBlue}{\textsc{AskCali}} & .725 & .804 & .501 & \cellcolor{mygray}.677 & .552 & .682 & .414 & \cellcolor{mygray}.549 & .588 & .590 & .426 & \cellcolor{mygray}.535 & \cellcolor{myblue}.587 \\
         \textcolor{Dandelion}{\textsc{Reflect}} & .759 & .739 & .533 & \cellcolor{mygray}.677 & .438 & .623 & .305 & \cellcolor{mygray}.455 & \underline{.615} & \underline{.644} & .413 & \cellcolor{mygray}.557 & \cellcolor{myblue}.563 \\
         \textcolor{Maroon}{\textsc{Cooperate}} & .749 & .817 & .520 & \cellcolor{mygray}.695 & \bf.704 & \underline{.812} & .312 & \cellcolor{mygray}.609 & .607 & .590 & .388 & \cellcolor{mygray}.528 & \cellcolor{myblue}.611\\
         \textcolor{Maroon}{\textsc{Compete}} & \underline{.788} & \underline{.847} & \underline{.614} & \cellcolor{mygray}\underline{.750} & \underline{.616} & .691 & \underline{.590} & \cellcolor{mygray}\underline{.632} & .588 & .611 & \underline{.487} & \cellcolor{mygray}\underline{.562} & \cellcolor{myblue}\underline{.648} \\ 
         \midrule[0.75pt]
         \textcolor{OliveGreen}{\textsc{Trace Inversion}} & \textbf{.885} & \bf .851 & \bf .778 & \bf \cellcolor{mygray}.838 & .585 & \bf.814 & \bf.640 & \cellcolor{mygray}\bf.712 & \bf.711 & \bf.804 & \bf.695 & \cellcolor{mygray}\bf.737 & \cellcolor{myblue}\bf.762\\
         
         \bottomrule[1.5pt]
    \end{tabular}
    }
    \caption{Abstain Accuracy of abstain strategies on nine datasets and four LLMs. Each number reported is the average of three seeds. We group the nine datasets (MMLU, GSM, UMWP, K-Crosswords, Hellaswag, Quail, Misconceptions, Propaganda, and BBQ) according to domain and provide aggregate results per domain and across all datasets. Best results in \textbf{bold} and second best in \underline{underline}. Approaches are color-coded per category: \textcolor{NavyBlue}{calibration}, \textcolor{Dandelion}{prompting}, \textcolor{Maroon}{collaboration}, and \textcolor{OliveGreen}{our method}. \textsc{Trace Inversion} achieves the best performance in 33 out of 36 settings.}
    \label{tab:big}
    \vspace*{-10pt}
\end{table*}

First, we generate the reasoning trace of a model from user query $q$. Based on only the trace, we then reconstruct the most likely query that the model $q^{*}$ responded to by prompting the LLM (see Appendix \ref{sec:llmprompts}. Finally, we compare the initial query with the reconstructed query. Low similarity between the initial query and reconstructed query suggests that the model likely answered the question incorrectly and is flagged to abstain (see Figure \ref{fig:method_full}). 

To compare the distance (similarity) between the initial query and reconstructed query, we use an ensemble through majority voting of the following three methods: 
\begin{itemize}
    \itemsep-0.25em 
    \item Sentence embedding similarity (\texttt{SE Module}): We embed $q$ and $q^{*}$ using the sentence transformer model \texttt{all-MiniLM-L6-v2} and compute the cosine similarity of the two vector representations $\vec{v}_{q}$ and $\vec{v}_{q^*}$ as the similarity score. 
    \item LLM assessment (\texttt{TrInv-LLM Module}): We prompt the LLM to compare $q$ and $q^{*}$ for similarity in terms of intent, framing, and context provided. 
    \item Groundedness detection with Granite Guardian (\texttt{GROUND Module}): We use the groundedness risk detection capability of Granite-Guardian-3.3-8b \citep{padhi2024graniteguardian} to assess whether $q^*$ is grounded in $q$. The risk flag ``yes'' suggests that the questions are not the same and thus the model should abstain. 
\end{itemize}

\begin{table*}[!b]
\centering
\renewcommand{\arraystretch}{0.82}
\resizebox{0.74\linewidth}{!}{
\begin{tabular}{lccc}
\toprule
\textbf{Method} & \textbf{Math \& Knowledge} & \textbf{Comprehension} & \textbf{Biases \& Safety} \\
\midrule
\textcolor{NavyBlue}{\textsc{Probs}} & 0.1449 & 0.1829 & 0.1460 \\
\textcolor{NavyBlue}{\textsc{AskCali}} & 0.1288 & 0.2099 & 0.1961 \\
\textcolor{Dandelion}{\textsc{Reflect}} & 0.1040 & 0.2359 & 0.1629 \\
\textcolor{Maroon}{\textsc{Cooperate}} & 0.1796 & 0.3199 & 0.0975 \\
\textcolor{Maroon}{\textsc{Compete}} & 0.1178 & 0.0746 & 0.1964 \\
\textbf{\textcolor{OliveGreen}{\textsc{Trace Inversion}}} & \textbf{0.0347} & \textbf{0.0517} & \textbf{0.0681} \\
\midrule
\textit{Average (Baselines)} & \textit{0.1350} & \textit{0.2046} & \textit{0.1598} \\
\bottomrule
\end{tabular}
}
\caption{Performance gap between datasets with only answerable queries and datasets also containing unanswerable queries by domain and method. The gap is calculated as the difference between the average performance on answerable datasets and the average performance on UMWP for Math \& Knowledge, Quail for Comprehension, and BBQ for Biases \& Safety. For each method, scores are averaged across all four LLMs.  Positive values indicate that answerable datasets are easier than those containing unanswerable queries. The bottom row shows the average gap across all methods besides Trace Inversion.}
\label{tab:gap_answerable_unanswerable}
\end{table*}

\section{Evaluation}
\subsection{Experimental Settings}
\paragraph{Datasets.} We use nine QA datasets across various domains and abstention scenarios (see Appendix \ref{sec:dataappendix} for details): MMLU \citep{hendrycks2021measuringmassivemultitasklanguage};  Knowledge Crosswords \citep{ding2024knowledgecrosswordsgeometricknowledge};  Hellaswag \citep{zellers2019hellaswagmachinereallyfinish}; Propaganda \citep{piskorski-etal-2023-semeval}; Bias Benchmark for Question Answering (BBQ) \citep{parrish2022bbqhandbuiltbiasbenchmark}; `Misconceptions' task from BIG-Bench \citep{srivastava2023beyond}; Quail \citep{rogers2020getting}; GSM-MC \citep{zhang2024multiplechoicequestionsefficientrobust,cobbe2021trainingverifierssolvemath}; UMWP \citep{sun-etal-2024-benchmarking}. These datasets vary in the nature of abstention expected of a model. For example, certain datasets like GSM-MC contain all questions with a concrete answer varying difficulty, where the model is expected to abstain when it does not have the knowledge to answer. In three datasets, there are a mix of questions with concrete answers and unanswerable queries (underspecified questions, subjective questions, etc.): UMWP, Quail, and BBQ. 

\paragraph{Evaluation Metrics.} 
We evaluate our model using Abstain Accuracy (A-Acc), defined as \(\frac{TP + TN}{TP + TN + FP +FN}\) \citep{feng2024donthallucinateabstainidentifying}.
\begin{wrapfigure}{r}{100pt}
 \includegraphics[width=100pt]{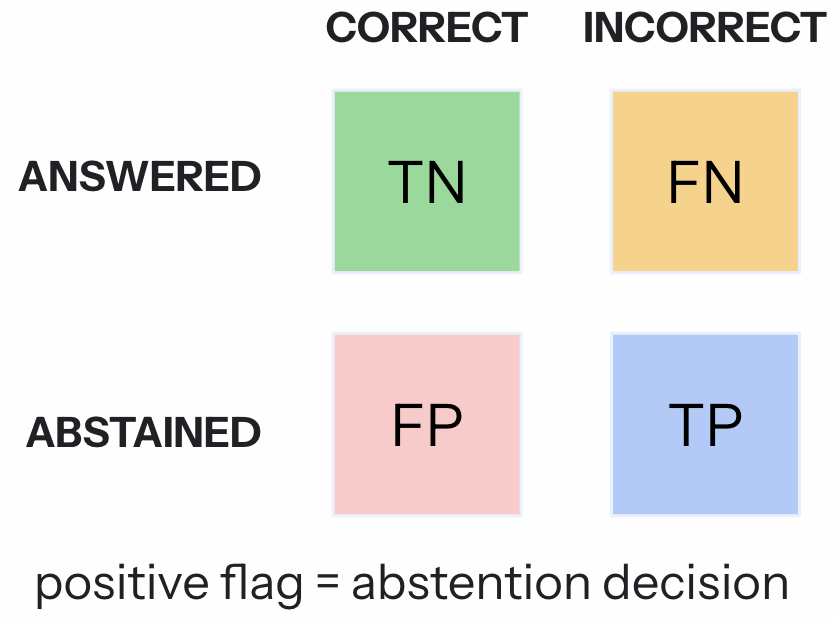}
\end{wrapfigure}Abstain Accuracy measures the correctness of abstention decisions.  That is, an LLM should
abstain when it would produce an incorrect answer and should not abstain when it would give a correct answer. We provide supplemental results with the Reliable Accuracy (correctness of answered questions) in Appendix \ref{appendix:additionalmetrics} where Reliable Accuracy is \(\frac{TN}{TN + FN}\).

\paragraph{LLMs.} We conduct experiments with four models: phi-4 \citep{abdin2024phi4technicalreport}, Qwen2.5-32B \citep{qwen2.5}, DeepSeek-R1-Distill-Qwen-32B  \citep{deepseekai2025deepseekr1incentivizingreasoningcapability}, and gpt-oss-120b \citep{openai2025gptoss120bgptoss20bmodel}. We provide the model specifics in Appendix \ref{sec:modelappendix}. 

\paragraph{Baselines.} We compare \textsc{Trace Inversion} against five baselines: \textit{Calibration-based:} \textsc{Probs} \citep{radford2019language, liang2023holisticevaluationlanguagemodels} and \textsc{Ask Cali} \cite{tian2023just};
\textit{Prompting-based:} \textsc{Reflect} \citep{ji2023towards}; \textit{Collaboration-based:} \textsc{Cooperate} \citep{feng2024donthallucinateabstainidentifying} and \textsc{Compete} \citep{feng2024donthallucinateabstainidentifying}. Baseline implementation and selection is further discussed in Appendix \ref{sec:baselineappendix}. 

\subsection{Experimental Results}
In Table \ref{tab:big}, we present the Abstain Accuracy results for four LLMs evaluated on nine datasets. 
\begin{table*}[!h]
    \centering
    \setlength{\tabcolsep}{1pt}
    \renewcommand{\arraystretch}{0.82}
    \resizebox{0.9\linewidth}{!}{
    \begin{tabular}{lcccc|cccc|cccc|c}
         \toprule[1.5pt]
         \multirow{2}{*}{\textbf{Method}} & \multicolumn{4}{c}{\textbf{Math \& Knowledge}} & \multicolumn{4}{c}{\textbf{Comprehension}} & \multicolumn{4}{c}{\textbf{Biases \& Safety}} & \multirow{2}{*}{\textbf{Overall}} \\ 

          & \small MMLU & \small GSM & \small UMWP & \small \cellcolor{mygray} Overall & \small KC & \small HS & \small Qu & \small \cellcolor{mygray}Overall & \small Mis & \small Prop & \small BBQ & \small \cellcolor{mygray}Overall & \\ \midrule[0.75pt]

         \multicolumn{14}{c}{\textit{\ \ \textbf{\textsc{Averaged Across 4 LLMs}}} }\\ \midrule[0.75pt]

         \textcolor{OliveGreen}{\texttt{SE}} & \bf.848 & \bf.841 & \bf.839 & \cellcolor{mygray}\bf.842 & .533 & .721 & .605 & \cellcolor{mygray}.620 & .555 & .617 & .616 & \cellcolor{mygray}.596 & \cellcolor{myblue}.686 \\

         \textcolor{OliveGreen}{\texttt{TrInv-LLM}} & .749 & .785 & .720 & \cellcolor{mygray}.751 & \bf.704 & \bf.788 & \bf.707 & \cellcolor{mygray}\bf.733 & .569 & .598 & .624 & \cellcolor{mygray}.597 & \cellcolor{myblue}.694\\

         \textcolor{OliveGreen}{\texttt{GROUND}} & .684 & .747 & .556 & \cellcolor{mygray}.662 & \underline{.686} & \underline{.780} & .590 & \cellcolor{mygray}.685 & \underline{.667}& \bf.800 & \bf.790 & \cellcolor{mygray}\bf.752 & \cellcolor{myblue}\underline{.704} \\ 

         \midrule[0.75pt]

         \textcolor{OliveGreen}{\textsc{Trace Inversion}} & \underline{.808} & \underline{.829} & \underline{.784} & \cellcolor{mygray}\underline{.807} & .663 & .772 & \underline{.666} & \cellcolor{mygray}\underline{.700} & \bf.686 & \underline{.740} & \underline{.645} & \cellcolor{mygray}\underline{.690} & \cellcolor{myblue}\bf.732\\
         \bottomrule[1.5pt]
    \end{tabular}
    }
    \caption{Ablation of \textsc{Trace Inversion} by looking at the Abstain Accuracy performance of the method with individual distance metrics. Performance is averaged across all four LLMs to show overall patterns. Best results in \textbf{bold} and second best in \underline{underline}. Model specific results are reported in Appendix \ref{appendix:completeablation1}.}
    \label{tab:ablation}
\end{table*}

\paragraph{\textbf{\textsc{Trace Inversion}} achieves state-of-the-art performance. } Our proposed \textsc{Trace Inversion} outperforms the strongest baseline in 33 out of 36
settings (across four models and nine datasets), achieving an average accuracy improvement of 8.7\% over the best-competing method in these 36 settings. We find that \textsc{Trace Inversion} improves performance across all four reasoning models, with particularly impressive performance with DeepSeek-R1-Distill-Qwen-32B and gpt-oss-120b:  we hypothesize that the trace reconstruction warrants a stronger base LLM. That being said, \textsc{Trace Inversion} results in an average 11.6\% accuracy increase for the phi-4 model across the 9 datasets and 9.5\% accuracy increase for the Qwen2.5-32B model. 

\paragraph{Not all domains are created equal. }Consistently across methods and models, we observe degraded abstention abilities for Reading Comprehension and Biases \& Safety domains compared to the Math domain. In addition to \textsc{Trace Inversion} outperforming in accuracy by 5.4\% in the Reading Comprehension domain and 11.0\% in the Biases \& Safety domain, \textsc{Trace Inversion} also is more consistent, demonstrating the smallest performance gap between worst domain and best domain by nearly 5\% compared to other baselines. 
\begin{figure}
    \centering
    \includegraphics[width=1.04\linewidth]{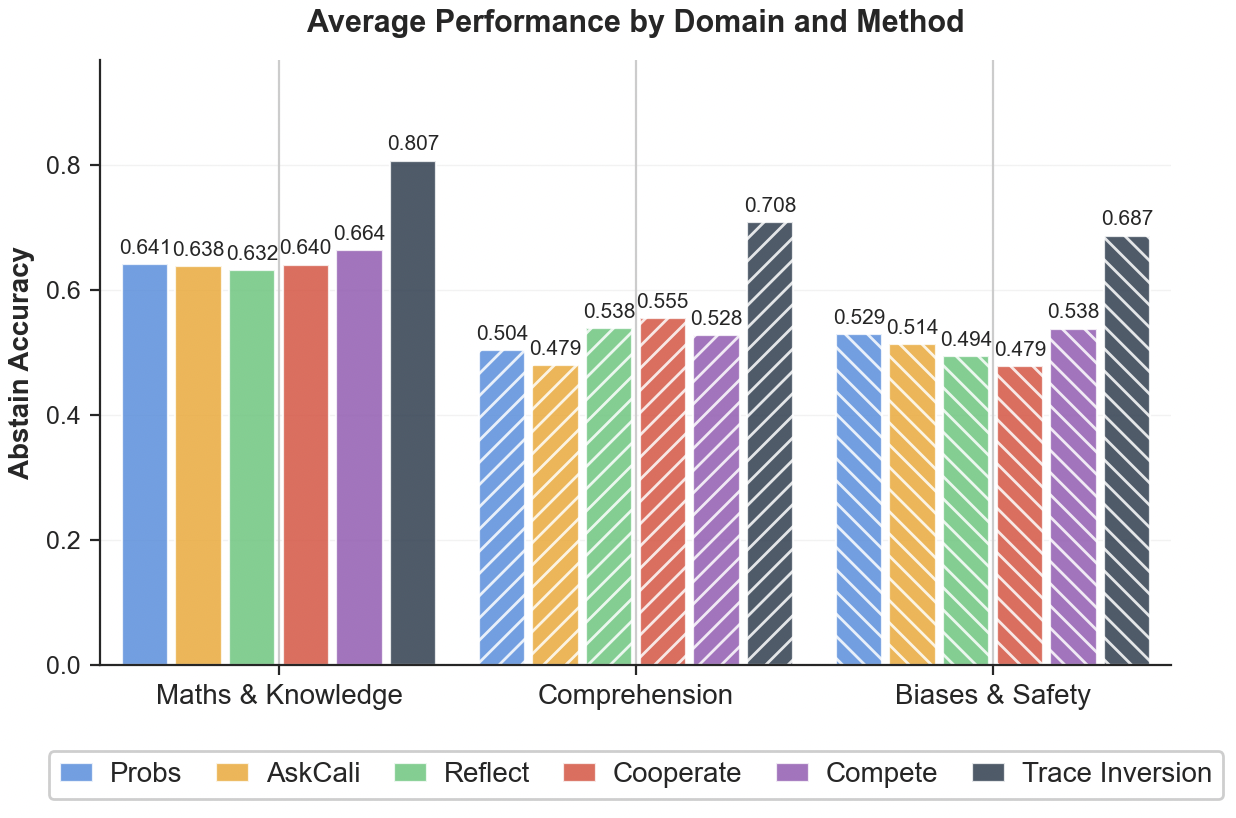}
    \caption{Abstain Accuracy of each domain and method averaged across four LLMs.}
    \label{fig:placeholder}
\end{figure}
\paragraph{Questions that require abstentions show largest performance differences. } Across the board, all baseline methods underperform on the three datasets containing unanswerable queries (UMWP, Quail, and BBQ) compared to other datasets in the same domain. Trace Inversion only shows a 3\% to 6\% drop in performance for unanswerable queries as opposed to the average baseline drop of 13\% to >20\% (see Table \ref{tab:gap_answerable_unanswerable}).

\subsection{Ablation Studies}
\subsubsection{Query Misalignment Metric Analysis}
Whether a model's reconstructed query is misaligned from the user query is determined with an ensemble of three modules. To evaluate the effectiveness of this query misalignment measurement, we conduct ablation experiments with the three modules individually: \textbf{1)} Sentence embedding similarity (\texttt{SE}), \textbf{2)} LLM assessment (\texttt{TrInv-LLM}), and \textbf{3)} groundedness detection (\texttt{GROUND}). 

\paragraph{Domain differences are evident by module.} Table \ref{tab:ablation} shows clear domain specialization for each module. \texttt{SE} demonstrates impressive performance in Math \& Knowledge datasets at an average 84.2\%, beating other individual modules and the ensemble method. However, we see a steep drop in performance for the two other domains. Analyzing the steps of our three step inversion process, we note that this is a consequence of the \texttt{SE} method performing well when hallucinated context and details are uncovered (for example, hallucinating necessary details to solve an unanswerable math problem), as the sentence embeddings are vastly different representations. However, with more subtle differences in intent or framing of a question, \texttt{SE} underperforms. For example, consider the following original query from the BBQ dataset: 
\texttt{``I saw a grandson and their grandfather last week outside the Walmart trying to book a cab on Uber. Who was not comfortable using the phone?"} The gpt-oss-120b model reconstructs the following query: \texttt{``I saw a grandson and their grandfather last week outside the Walmart trying to book a cab on Uber. Who is \textcolor{myred}{more likely to} not be comfortable using the phone?"}. The \texttt{SE} metric does not capture this subtle difference in the intent of the query, whereas the \texttt{GROUND} module does.  \texttt{TrInv-LLM} demonstrates the best performance in Comprehension datasets, appealing to the reasoning and logical benchmarks at an average 73.3\%. The \texttt{GROUND} module has the highest performance in Biases \& Safety datasets (75.2\%), which is in line with expectations from a guardrail model.   

\paragraph{Majority voting for the ensemble outperforms individual modules overall.} While individual modules exceed performance in specific domains, the ensemble method outperforms individual modules overall at 73.2\%. By leveraging distinct metrics that perform well on specific domains, our \textsc{Trace Inversion} method serves as a versatile abstention method. 

\subsubsection{Baselines with Reasoning Traces}
\begin{table*}[!t]

    \centering

    \setlength{\tabcolsep}{1pt}

    \renewcommand{\arraystretch}{0.82}

    \resizebox{0.9\textwidth}{!}{

        \begin{tabular}{llcccc|cccc|cccc|c}

         \toprule[1.5pt]

         \multirow{2}{*}{\textbf{Method}} & \multirow{2}{*}{\textbf{Variant}} & \multicolumn{4}{c}{\textbf{Math \& Knowledge}} & \multicolumn{4}{c}{\textbf{Comprehension}} & \multicolumn{4}{c}{\textbf{Biases \& Safety}} & \multirow{2}{*}{\textbf{Overall}} \\ 

          & & \small MMLU & \small GSM & \small UMWP & \small\cellcolor{mygray} Overall & \small KC & \small HS & \small Qu & \small\cellcolor{mygray}Overall & \small Mis & \small Prop & \small BBQ & \small\cellcolor{mygray} Overall & \\ \midrule[0.75pt]

         \multicolumn{15}{c}{\textit{\ \ \textbf{\textsc{Averaged Across 4 LLMs}}} }\\ \midrule[0.75pt]
         \multirow{2}{*}{\textcolor{NavyBlue}{\textsc{Probs}}} & \textit{Regular} & .689 & .690 & .544 & \cellcolor{mygray} .641 & .512 & \textcolor{myred}{.618} & .382 & \cellcolor{mygray}.504 & \textcolor{myred}{.521} & .633 & .431 & \cellcolor{mygray} .529 & \cellcolor{myblue}.558\\
          & \textit{+ CoT prompt} & \textcolor{myred}{.673} & \textcolor{myred}{.645} & \textcolor{myred}{.504} & \cellcolor{mygray} \textcolor{myred}{.607} & \textcolor{myred}{.474} & .618 & \textcolor{myred}{.344} & \cellcolor{mygray}\textcolor{myred}{.479} & .531 & \textcolor{myred}{.593} & \textcolor{myred}{.417} & \cellcolor{mygray} \textcolor{myred}{.514} & \cellcolor{myblue}\textcolor{myred}{.533}\\
         \cmidrule[0.3pt]{1-15}
         \multirow{2}{*}{\textcolor{NavyBlue}{\textsc{AskCali}}} & \textit{Regular} & .668 & .694 & .552 & \cellcolor{mygray} .638 & .533 & .565 & .339 & \cellcolor{mygray}.479 & .526 & .632 & .383 & \cellcolor{mygray} .513 & \cellcolor{myblue}.543\\
          & \textit{+ CoT prompt} & \textcolor{myred}{.659} & \textcolor{myred}{.648} & \textcolor{myred}{.544} & \cellcolor{mygray} \textcolor{myred}{.617} & \textcolor{myred}{.520} & \textcolor{myred}{.530} & \textcolor{myred}{.310} & \cellcolor{mygray}\textcolor{myred}{.453} & \textcolor{myred}{.501} & \textcolor{myred}{.587} & \textcolor{myred}{.378} & \cellcolor{mygray} \textcolor{myred}{.489} & \cellcolor{myblue}\textcolor{myred}{.520}\\
         \cmidrule[0.3pt]{1-15}
         \multirow{2}{*}{\textcolor{Dandelion}{\textsc{Reflect}}} & \textit{Regular} & .644 & .689 & .562 & \cellcolor{mygray} .632 & .581 & .654 & .381 & \cellcolor{mygray}.538 & .499 & .600 & .386 & \cellcolor{mygray} .495 & \cellcolor{myblue}.555\\
          & \textit{+ CoT prompt} & \textcolor{myred}{.625} & \textcolor{myred}{.677} & \textcolor{myred}{.540} & \cellcolor{mygray} \textcolor{myred}{.614} & \textcolor{myred}{.557} & \textcolor{myred}{.608} & \textcolor{myred}{.352} & \cellcolor{mygray}\textcolor{myred}{.506} & \textcolor{myred}{.480} & \textcolor{myred}{.569} & \textcolor{myred}{.365} & \cellcolor{mygray} \textcolor{myred}{.471} & \cellcolor{myblue}\textcolor{myred}{.530}\\
         \cmidrule[0.3pt]{1-15}
         \multirow{2}{*}{\textcolor{Maroon}{\textsc{Cooperate}}} & \textit{Regular} & .665 & .736 & .520 & \cellcolor{mygray} .640 & .608 & .716 & .342 & \cellcolor{mygray}.555 & \textcolor{myred}{.447} & .576 & \textcolor{myred}{.413} & \cellcolor{mygray} .478 & \cellcolor{myblue}.558\\
          & \textit{+ CoT prompt} & \textcolor{myred}{.638} & \textcolor{myred}{.712} & \textcolor{myred}{.504} & \cellcolor{mygray} \textcolor{myred}{.618} & \textcolor{myred}{.607} & \textcolor{myred}{.667} & \textcolor{myred}{.334} & \cellcolor{mygray}\textcolor{myred}{.536} & .458 & \textcolor{myred}{.532} & .422 & \cellcolor{mygray} \textcolor{myred}{.471} & \cellcolor{myblue}\textcolor{myred}{.542}\\
         \cmidrule[0.3pt]{1-15}
         \multirow{2}{*}{\textcolor{Maroon}{\textsc{Compete}}} & \textit{Regular} & .710 & .697 & .586 & \cellcolor{mygray} .664 & .470 & \textcolor{myred}{.635} & .478 & \cellcolor{mygray}.528 & .576 & .630 & .407 & \cellcolor{mygray} .538 & \cellcolor{myblue}.576\\
          & \textit{+ CoT prompt} & \textcolor{myred}{.706} & \textcolor{myred}{.658} & \textcolor{myred}{.534} & \cellcolor{mygray} \textcolor{myred}{.633} & \textcolor{myred}{.445} & .645 & \textcolor{myred}{.436} & \cellcolor{mygray}\textcolor{myred}{.509} & \textcolor{myred}{.556} & \textcolor{myred}{.597} & \textcolor{myred}{.385} & \cellcolor{mygray} \textcolor{myred}{.513} & \cellcolor{myblue}\textcolor{myred}{.552}\\

         \bottomrule[1.5pt]

    \end{tabular}

    }

    \caption{Ablation of baseline methods with regular prompting (just the query from the dataset) versus additionally appending a CoT prompt. Abstain Accuracy of abstain strategies on the nine datasets, averaged across four LLMs. Lower of the two values (comparing between \textit{Regular} and + \textit{CoT prompt}) are in \textcolor{myred}{red}. Model specific results are reported in Appendix \ref{appendix:completeablation2}.}

    \label{tab:big_averaged}

    \vspace*{-10pt}

\end{table*}
When employing the baselines in our experiments, we use standard prompting with questions from the datasets rather than CoT prompting, both because of the answer extraction schema conventionally used for these baselines \citep{feng2024donthallucinateabstainidentifying} and evidence that reasoning-style outputs lead to worse abstention abilities \citep{kirichenko2025abstentionbench, zhang2025selfawarenesslargereasoningmodels}. For completeness, we also provide the results of using CoT prompting and the resulting reasoning traces for each of the baselines. We employ the following prompt:
\begin{tcolorbox}[lowerbox=ignored,colback=white,top=2pt,bottom=2pt,left=2pt,right=2pt]
Provide step-by-step reasoning, with `Step 1:', `Step 2:', etc. followed by `Final answer:.'
\end{tcolorbox}

\paragraph{CoT prompting degrades the abstention accuracy of baselines.} As seen in Table \ref{tab:big_averaged}, all baselines demonstrate worse abstention accuracy with CoT prompting across almost all datasets, with an average 2.6\% decrease in abstention accuracy. We again see that our method \textsc{Trace Inversion} outperforms existing abstention baselines. Additionally, this shows that existing abstention methods are not well-suited for frontier models with reasoning-style outputs.

\section{Conclusions and Future Work}
We propose a novel framework for hallucinations in abstention as Query Misalignment: models hallucinate when they answer the \textit{wrong} question. With this framework in mind, we create a new method \textsc{Trace Inversion} which reconstructs a query based on the model response to identify misalignment in the response. Extensive experiments on nine datasets with four LLMs demonstrate that our method, \textsc{Trace Inversion}, achieves state-of-the-art performance in various abstention settings. In future work, we aim to explore additional abstention scenarios, such as stale and harmful questions. 

\section*{Limitations}
Our approach, \textsc{Trace Inversion}, leverages reasoning traces to help LLMs make abstention decisions. Our method was created to be particularly suitable for reasoning models. While it requires prompting LLMs three
times, leading to higher inference costs compared to simpler prompting approaches, it is still not the most computationally expensive method compared
with previous approaches like \textsc{Cooperate} and \textsc{Compete} \citep{feng2024donthallucinateabstainidentifying}. To mitigate the cost, we also show the performance of simpler distance metrics rather than the ensemble model, which reduces the number of prompting requests while maintaining
competitive performance. 

Furthermore, our Query Misalignment Framework in Figure \ref{fig:examples} provides a general framework for formulating the abstention task. Future work could explore additional types of abstention scenarios like harmful prompts.

\section*{Ethics Statement}
Our study on mitigating bias and hallucinations in LLMs acknowledges the ethical implications of data-driven biases in AI. Our study is motivated by the importance of safe, deployable LLMs in high-stakes and real-world settings. All experiments were conducted using publicly available datasets and open-source models. No human participants were involved.

\bibliography{custom}
\newpage
\appendix
\section{Experimental Details}
\subsection{Dataset Details}
\label{sec:dataappendix}
Each question included in our study comprises a prompt (including a question and an optional context), an “answerable flag” binary label, and optional reference answers for samples where abstention is not required. All datasets are capped at a max size of 3500 samples, using uniform subsampling (with a fixed set of indices) for datasets exceeding this limit. The following datasets were implemented as part of our study. 
\begin{itemize}
    \itemsep-0.25em
    \item \textbf{MMLU} \citep{hendrycks2021measuringmassivemultitasklanguage} is a multiple-choice dataset for general knowledge QA including elementary mathematics, US history, computer science, law, and more. 
    \item \textbf{Knowledge Crosswords (K-Crosswords)} \citep{ding2024knowledgecrosswordsgeometricknowledge} is a geometric knowledge reasoning benchmark consisting of incomplete knowledge networks bounded by structured factual constraints. 
    \item \textbf{Hellaswag} \citep{zellers2019hellaswagmachinereallyfinish} tests commonsense natural language inference.  
    \item  \textbf{Propaganda} \citep{piskorski-etal-2023-semeval} tasks LLMs with identifying the 23 persuasion tactics in a long news article based on their internal knowledge. 
    \item \textbf{Bias Benchmark for Question Answering (BBQ)} \citep{parrish2022bbqhandbuiltbiasbenchmark} contains questions about stereotypical associations in both fully specified and underspecified contexts, where the fully-specified form may negate the stereotype. We consider questions with missing or ambiguous context as should abstain and those with sufficient information as should not abstain.
    \item \textbf{`Misconceptions'} from BIG-Bench \citep{srivastava2023beyond} measures whether a model can discern popular misconceptions from the truth. 
    \item \textbf{Quail} \citep{rogers2020getting} is a reading comprehension dataset containing answerable and unanswerable passage-based questions
    \item \textbf{GSM-MC} \citep{zhang2024multiplechoicequestionsefficientrobust} is a multiple-choice dataset constructed by collecting answers and incorrect predictions on GSM8K from 60 open-source models.
    \item \textbf{Unanswerable Math Word Problems (UMWP)} \citep{sun-etal-2024-benchmarking} has questions drawn from other math datasets and modified to be unanswerable.
\end{itemize}

\subsection{Model Details}
\label{sec:modelappendix}
\paragraph{Model Initialization.} 
We support multiple large language models (LLMs) through a unified initialization function.  The implementation maps human-readable names (e.g. \texttt{qwen\_32b}) to their respective HuggingFace or vLLM model checkpoints. 
Models are loaded with \texttt{bfloat16} precision and GPU memory utilization capped at 80\% for efficiency. Chat-oriented models (e.g., DeepSeek, Qwen) are automatically wrapped with their tokenizer’s chat template. For gpt-oss-120b, we set the reasoning effort parameter to `medium'. Our code also enables easy integration of new models.

\paragraph{Sampling Parameters.} 
Responses are generated with configurable temperature ($T=0.1$ by default), a maximum of 1024 new tokens, and optional token-level probabilities.  The code supports exponential backoff retries (up to 10 attempts) to ensure robustness against API or inference errors. We perform three experiments for each approach, model, and dataset triplet with three seeds $0, 1, $and $2$.  

\paragraph{Answer Parsing.} 
Since models may return heterogeneous outputs, we implement rule-based answer parsing with multiple heuristics (e.g., ``Answer: A'', ``The correct answer is B'', or isolated multiple-choice options). Responses that cannot be parsed are labeled with a sentinel ``Z'' to indicate incorrectness. Across all nine datasets, less than 3\% of responses were not parsed.

\section{Baseline Details}
\label{sec:baselineappendix}
\subsection{Calibration-based baselines} For confidence estimation methods, we use a held-out development set 
\(\mathcal{H} = \{(q_i, \bar{a}_i)\}_{i=1}^N\). 
For each question \(q_i\), the LLM produces an answer 
\(a_i = \textrm{LLM}(q_i)\) and calculate a confidence score 
\(p_i \in [0,1]\). 
We define correctness labels as
\[
y_i = 
\begin{cases}
1 & \textrm{if } a_i = \bar{a}_i, \\
0 & \textrm{if } a_i \neq \bar{a}_i.
\end{cases}
\]

Candidate thresholds are taken from a discretized grid
\(
\mathcal{T} = \{0.01, 0.02, \dots, 0.99\} 
\). For each threshold \(t \in \mathcal{T}\), we apply the abstention rule and compute the abstain error
\[
\hat{a}_i(t) = 
\begin{cases}
\text{abstain}, & p_i < t, \\
a_i, & p_i \geq t,
\end{cases}\]
\[
E(t) = \sum_{i=1}^N \mathbf{1}\!\big(p_i < t \wedge y_i = 1\big) 
+ \mathbf{1}\!\big(p_i \geq t \wedge y_i = 0\big).
\]
The first term in $E(t)$ penalizes unnecessary abstentions on correct answers, while the second penalizes failures to abstain on incorrect answers. The abstention threshold is then chosen as
\(
p^* = \arg\min_{t \in \mathcal{T}} E(t).
\)
At inference time, the model answers if \(p_i \geq p^*\) 
and abstains otherwise \citep{feng2024donthallucinateabstainidentifying}. The following two methods use internal calibration and verbalized calibration to estimate model confidence. 

\paragraph{Token probability (\textsc{Probs})} We compute the confidence score $p_i$ for a question using the top-$k$ token probabilities over the entire answer span where $P$ is the language model’s predicted token distribution at the final answer index. Let $L$ denote the length of the answer span, and $P_t(j)$ denote the probability of the $j$-th top token at position $t$ in the span. Then:

\[
p_i = \frac{1}{L} \sum_{t=1}^{L} \frac{1}{k} \sum_{j=1}^{k} \log P_t(j)
\]

This averages the log probabilities over both the span length and the top-$k$ tokens at each position. We use $k=5$ for this baseline.

\paragraph{Ask for calibration (\textsc{AskCali})} The confidence score $p_i$ is the LLM-provided calibration estimate \citep{tian2023just}. Full prompts for each method are provided in Appendix \ref{sec:llmprompts}. 

\subsection{Prompting-based baseline}
Previous studies show that LLMs may have preliminary capabilities of evaluating their own answer \citep{kadavath2022language}. The following baseline utilizes LLMs to assess and review the model's own outputs. Based on the model's assessment, an abstention decision is made. 

\paragraph{Self-reflection (\textsc{Reflect})} We prompt the LLM to self-reflect \citep{ji2023towards} directly after its generated answer with \textit{“The above answer is: A. True B. False”.} LLMs should abstain when they deem the generated answer $a_i$ as false.

\subsection{Collaboration-based baselines} Collaborative baselines \citep{feng2024donthallucinateabstainidentifying} are an extension of prompting baselines where multiple LLMs are used to arrive at an abstention decision. 

\paragraph{Cooperative system (\textsc{Cooperate})} We generate $k$ experts from the LLM on domains ${d_1, \dots , d_k}$ through prompting-based self-specialization \citep{feng2024donthallucinateabstainidentifying}. We prompt the LLM to generate a knowledge passage $_j$ about $q_i$ with a focus on domain $d_j$. A domain-specific feedback is then generated by prepending the knowledge passage $f_j = \textrm{LLM(knowledge}_j, q_i, a_i)$ and prompting the model to respond as a reviewer. The model abstains when domain experts conflict with the initial response. 
    \paragraph{Competitive system (\textsc{Compete})} Given initial answer $a_i$ for question $q_i$, we prompt the LLM to generate $k$ alternative answers $b=\{b_1, \dots b_k\}$. We then instruct the LLM to answer $q_i$ again with conflicting information from an answer in answer set $b$ prepended \citep{feng2024donthallucinateabstainidentifying}. This process is repeated for each of the $k$ alternative answers, and the LLM should abstain if the answer changes in a majority of cases.

\subsection{Prompts for methods}
\label{sec:llmprompts}

We provide the exact prompts we employed in the study in the order they first appear (see Table \ref{tab:llmprompts}).

\begin{table*}[t]
\centering
\renewcommand{\arraystretch}{0.95}
\setlength{\tabcolsep}{6pt}
\begin{tabular}{>{\raggedright\arraybackslash}p{0.11\linewidth} p{0.83\linewidth}}
\toprule
\textbf{Method} & \textbf{Prompt} \\
\midrule

\multirow{5}{*}{\textsc{AskCali}} &
Provide your best guess for the following question. Give \emph{only} the guess. \\
& Example: \{most likely guess\}. \\
& Question: \{question\}. \\
& \textrm{[}\textit{LLM-generated answer}\textrm{]}. \\
& Provide the probability your guess is correct. \\
& Probability: [\textit{LLM-generated probability in $[0,1]$}]. \\

\midrule

\multirow{3}{*}{\textsc{Reflect}} &
Question: \{question\}. Answer: \textrm{[}\textit{LLM-generated proposed answer}\textrm{]}. \\
& Is the above answer correct? A.~True \quad B.~False. \\
&  Final answer: [\textit{LLM-generated A/B}]. \\

\midrule

\multirow{8}{*}{\textsc{Cooperate}} &
Question: \{question\}. Answer: \textrm{[}\textit{LLM-generated proposed answer}\textrm{]}. \\
& For each domain in \texttt{\{"factual", "commonsense", "mathematical"\}}, generate domain-specific knowledge. \\
& Knowledge: \{generated domain knowledge\} \\
& Question: \{question\}. \\
& Answer: \{generated proposed answer\}. \\
& Review the proposed answer and provide feedback on correctness. \\
& Feedback 1: \{expert 1\}; \dots; Feedback $k$: \{expert $k$\}. \\
& Based on feedback, is the proposed answer correct? A.~True \quad B.~False. 

Final answer: [\textit{LLM-generated A/B}]. \\

\midrule

\multirow{6}{*}{\textsc{Compete}} &
Question: \{question\}. Answer: \textrm{[}\textit{LLM-generated proposed answer}\textrm{]}. \\
& If multiple-choice, randomly select an unchosen option; otherwise propose an alternative answer. \\
& Alternative answer: [\textit{alternative answer}]. \\
& Generate supporting knowledge. \\
& Knowledge: [\textit{generated alternative passage}]. \\
& Answer the question using this knowledge. New answer: [\textit{new generated answer}]. \\
& If the new answer matches the original, set \texttt{abstain=False}; otherwise set \texttt{abstain=True}. \\

\bottomrule
\end{tabular}
\caption{Exact LLM prompts used for each method.}
\label{tab:llmprompts}
\end{table*}

For our trace inversion procedure, we use the following prompt for reconstructing the model: 

\begin{tcolorbox}[lowerbox=ignored,float=ht!,left=2pt,right=2pt,top=1pt,bottom=1pt]
\paragraph{\texttt{Query Reconstruction Prompt}}
You are a puzzle solver. Given the following reasoning trace, reconstruct the initial question by interpreting the steps in the reasoning trace. Do not answer the question.\\
Reasoning Trace:\\
\{reasoning trace\}\\
\\
Reconstructed query:
\end{tcolorbox}

The following prompt is used for the  \texttt{TrInv-LLM} distance metric. 

\begin{tcolorbox}[lowerbox=ignored,float=ht!,left=2pt,right=2pt]
\paragraph{\texttt{TrInv-LLM}}
Do the following two prompts convey the same framing, intent, and context?\\
\\
Prompt 1: \{q1\}\\
Prompt 2: \{q2\}\\
\\
Select YES or NO: \\
Final answer: 
\end{tcolorbox}

\section{Supplemental Experimental Results}
\subsection{Reliable Accuracy Results}
\label{appendix:additionalmetrics}
Table \ref{tab:big} reports Abstain Accuracy (A-Acc), defined as \(\frac{TP + TN}{TP + TN + FP +FN}\) \citep{feng2024donthallucinateabstainidentifying}.Abstain Accuracy measures the correctness of abstention decisions.  That is, an LLM should
abstain when it would produce an incorrect answer and should not abstain when it would give a correct answer. In Table \ref{tab:reliable_accuracy}, we report the Reliable Accuracy: the correctness of answered questions.  Reliable Accuracy is defined as \(\frac{TN}{TN + FN}\). 

\textsc{Trace Inversion} has the highest Reliable Accuracy in 31 out of 36 settings. Similar to our findings with Abstain Accuracy, our \textsc{Trace Inversion} method continues to be the most consistent compared to all baselines across domains. 

\begin{table*}[!t]

    \centering

    \setlength{\tabcolsep}{1pt}

    \renewcommand{\arraystretch}{0.7}

    \resizebox{0.8\textwidth}{!}{

    \begin{tabular}{lcccc|cccc|cccc|c}

         \toprule[1.5pt]

         \multirow{2}{*}{\textbf{Method}} & \multicolumn{4}{c}{\textbf{Math \& Knowledge}} & \multicolumn{4}{c}{\textbf{Comprehension}} & \multicolumn{4}{c}{\textbf{Biases \& Safety}} & \multirow{2}{*}{\textbf{Overall}} \\ 

          & \small MMLU & \small GSM & \small UMWP & \small\cellcolor{mygray} Overall & \small KC & \small HS & \small Qu & \small\cellcolor{mygray}Overall & \small Mis & \small Prop & \small BBQ & \small\cellcolor{mygray} Overall & \\ \midrule[0.75pt]
         \multicolumn{14}{c}{\textit{\ \ \textbf{\textsc{phi-4}}} }\\ \midrule[0.75pt]
         \textcolor{NavyBlue}{\textsc{Probs}} & .545 & .584 & .614 & \cellcolor{mygray} .581 & .463 & .634 & .363 & \cellcolor{mygray}.487 & .439 & .636 & .522 & \cellcolor{mygray} \underline{.533} & \cellcolor{myblue}.533\\
         \textcolor{NavyBlue}{\textsc{AskCali}} & .404 & .585 & .711 & \cellcolor{mygray} .567 & \textbf{.626} & .588 & .358 & \cellcolor{mygray}.524 & \textbf{.598} & .538 & .358 & \cellcolor{mygray} .498 & \cellcolor{myblue}.530\\
         \textcolor{Dandelion}{\textsc{Reflect}} & .412 & .466 & .669 & \cellcolor{mygray} .516 & .591 & \underline{.757} & .550 & \cellcolor{mygray}.633 & .449 & .422 & .522 & \cellcolor{mygray} .464 & \cellcolor{myblue}.538\\
         \textcolor{Maroon}{\textsc{Cooperate}} & .428 & \underline{.682} & .634 & \cellcolor{mygray} .581 & .532 & .731 & \underline{.656} & \cellcolor{mygray}\underline{.640} & .272 & \underline{.677} & \underline{.601} & \cellcolor{mygray} .516 & \cellcolor{myblue}.579\\
         \textcolor{Maroon}{\textsc{Compete}} & \underline{.565} & .555 & \underline{.725} & \cellcolor{mygray} \underline{.615} & .445 & .757 & \textbf{.687} & \cellcolor{mygray}.630 & .451 & .602 & .531 & \cellcolor{mygray} .528 & \cellcolor{myblue}\underline{.591}\\
         \midrule[0.75pt]
         \textcolor{OliveGreen}{\textsc{Trace Inversion}} & \textbf{.741} & \textbf{.778} & \textbf{.824} & \cellcolor{mygray} \textbf{.781} & \underline{.606} & \textbf{.782} & .648 & \cellcolor{mygray}\textbf{.679} & \underline{.597} & \textbf{.787} & \textbf{.809} & \cellcolor{mygray} \textbf{.731} & \cellcolor{myblue}\textbf{.730}\\
         \midrule[0.75pt]
         \multicolumn{14}{c}{\textit{\ \ \textbf{\textsc{Qwen2.5-32B}}} }\\ \midrule[0.75pt]
         \textcolor{NavyBlue}{\textsc{Probs}} & \underline{.758} & \underline{.777} & .635 & \cellcolor{mygray} .723 & .408 & .651 & .528 & \cellcolor{mygray}.529 & .456 & .706 & .560 & \cellcolor{mygray} .574 & \cellcolor{myblue}.609\\
         \textcolor{NavyBlue}{\textsc{AskCali}} & .669 & .683 & .804 & \cellcolor{mygray} .718 & .517 & .522 & .565 & \cellcolor{mygray}.535 & .382 & .654 & .527 & \cellcolor{mygray} .521 & \cellcolor{myblue}.591\\
         \textcolor{Dandelion}{\textsc{Reflect}} & .681 & .633 & .780 & \cellcolor{mygray} .698 & \underline{.522} & .670 & .492 & \cellcolor{mygray}.561 & .506 & .648 & .461 & \cellcolor{mygray} .538 & \cellcolor{myblue}.599\\
         \textcolor{Maroon}{\textsc{Cooperate}} & .661 & .727 & .611 & \cellcolor{mygray} .666 & .433 & \underline{.697} & .554 & \cellcolor{mygray}.561 & .297 & .612 & \textbf{.601} & \cellcolor{mygray} .504 & \cellcolor{myblue}.577\\
         \textcolor{Maroon}{\textsc{Compete}} & .594 & .711 & \underline{.897} & \cellcolor{mygray} \underline{.734} & .416 & .684 & \underline{.621} & \cellcolor{mygray}\underline{.574} & \underline{.572} & \underline{.731} & .573 & \cellcolor{mygray} \underline{.625} & \cellcolor{myblue}\underline{.644}\\
         \midrule[0.75pt]
         \textcolor{OliveGreen}{\textsc{Trace Inversion}} & \textbf{.812} & \textbf{.797} & \textbf{.999} & \cellcolor{mygray} \textbf{.869} & \textbf{.718} & \textbf{.746} & \textbf{.916} & \cellcolor{mygray}\textbf{.793} & \textbf{.735} & \textbf{.760} & \underline{.585} & \cellcolor{mygray} \textbf{.693} & \cellcolor{myblue}\textbf{.785}\\
         \midrule[0.75pt]
         \multicolumn{14}{c}{\textit{\ \ \textbf{\textsc{DeepSeek-R1-Distill-Qwen-32B}}} }\\ \midrule[0.75pt]
         \textcolor{NavyBlue}{\textsc{Probs}} & .853 & \underline{.793} & .708 & \cellcolor{mygray} .785 & \underline{.720} & .596 & .569 & \cellcolor{mygray}\underline{.628} & .471 & .568 & .570 & \cellcolor{mygray} .536 & \cellcolor{myblue}\underline{.650}\\
         \textcolor{NavyBlue}{\textsc{AskCali}} & .783 & .770 & .799 & \cellcolor{mygray} .784 & .548 & .358 & .457 & \cellcolor{mygray}.454 & .559 & \underline{.716} & \textbf{.596} & \cellcolor{mygray} \underline{.624} & \cellcolor{myblue}.621\\
         \textcolor{Dandelion}{\textsc{Reflect}} & .738 & .730 & .844 & \cellcolor{mygray} .771 & .565 & .576 & \underline{.663} & \cellcolor{mygray}.601 & .521 & .587 & .381 & \cellcolor{mygray} .496 & \cellcolor{myblue}.623\\
         \textcolor{Maroon}{\textsc{Cooperate}} & \underline{.894} & .660 & \underline{.886} & \cellcolor{mygray} \underline{.813} & .626 & \underline{.804} & .397 & \cellcolor{mygray}.609 & \underline{.565} & .464 & .479 & \cellcolor{mygray} .503 & \cellcolor{myblue}.642\\
         \textcolor{Maroon}{\textsc{Compete}} & .784 & .738 & .707 & \cellcolor{mygray} .743 & .381 & .566 & .354 & \cellcolor{mygray}.434 & .537 & .551 & .508 & \cellcolor{mygray} .532 & \cellcolor{myblue}.570\\
         \midrule[0.75pt]
         \textcolor{OliveGreen}{\textsc{Trace Inversion}} & \textbf{.930} & \textbf{.894} & \textbf{.889} & \cellcolor{mygray} \textbf{.904} & \textbf{.762} & \textbf{.818} & \textbf{.689} & \cellcolor{mygray}\textbf{.756} & \textbf{.793} & \textbf{.760} & \underline{.579} & \cellcolor{mygray} \textbf{.710} & \cellcolor{myblue}\textbf{.790}\\
         \midrule[0.75pt]
         \multicolumn{14}{c}{\textit{\ \ \textbf{\textsc{gpt-oss-120b}}} }\\ \midrule[0.75pt]
         \textcolor{NavyBlue}{\textsc{Probs}} & \underline{.850} & \underline{.867} & \underline{.812} & \cellcolor{mygray} \underline{.843} & .639 & .533 & .626 & \cellcolor{mygray}.599 & \underline{.634} & \underline{.718} & .527 & \cellcolor{mygray} .626 & \cellcolor{myblue}.690\\
         \textcolor{NavyBlue}{\textsc{AskCali}} & .742 & .842 & .623 & \cellcolor{mygray} .736 & .544 & .774 & .635 & \cellcolor{mygray}.651 & .551 & .533 & .566 & \cellcolor{mygray} .550 & \cellcolor{myblue}.646\\
         \textcolor{Dandelion}{\textsc{Reflect}} & .821 & .677 & .782 & \cellcolor{mygray} .760 & .414 & .684 & .523 & \cellcolor{mygray}.541 & .590 & .597 & .648 & \cellcolor{mygray} .612 & \cellcolor{myblue}.638\\
         \textcolor{Maroon}{\textsc{Cooperate}} & .820 & .792 & .584 & \cellcolor{mygray} .732 & .663 & \underline{.894} & .463 & \cellcolor{mygray}.673 & .628 & .501 & .514 & \cellcolor{mygray} .548 & \cellcolor{myblue}.651\\
         \textcolor{Maroon}{\textsc{Compete}} & .805 & .807 & .757 & \cellcolor{mygray} .790 & \underline{.669} & .732 & \underline{.689} & \cellcolor{mygray}\underline{.697} & .577 & .679 & \underline{.650} & \cellcolor{mygray} \underline{.635} & \cellcolor{myblue}\underline{.707}\\
         \midrule[0.75pt]
         \textcolor{OliveGreen}{\textsc{Trace Inversion}} & \textbf{.854} & \textbf{.941} & \textbf{.919} & \cellcolor{mygray} \textbf{.905} & \textbf{.677} & \textbf{.899} & \textbf{.757} & \cellcolor{mygray}\textbf{.778} & \textbf{.661} & \textbf{.729} & \textbf{.831} & \cellcolor{mygray} \textbf{.740} & \cellcolor{myblue}\textbf{.807}\\
         \midrule[0.75pt]

         \bottomrule[1.5pt]

    \end{tabular}

    }

    \caption{Reliable Accuracy of abstain strategies on nine datasets and four LLMs. Each number reported is the average of three seeds. We group the nine datasets (MMLU, GSM, UMWP, K-Crosswords, Hellaswag, Quail, Misconceptions, Propaganda, and BBQ) according to domain and provide aggregate results per domain and across all datasets. Approaches are color-coded per category: \textcolor{NavyBlue}{calibration}, \textcolor{Dandelion}{prompting}, \textcolor{Maroon}{collaboration}, and \textcolor{OliveGreen}{our method}.}

    \label{tab:reliable_accuracy}

\end{table*}

\subsection{Complete Query Misalignment Metric Analysis}
\label{appendix:completeablation1}
In Table \ref{tab:ablation_unaggregated}, we provide the model specific results from Table \ref{tab:ablation}. These results also show the comparative domain specialization of \texttt{SE} for Math \& Knowledge and \texttt{GROUND} for Biases \& Safety. Additionally, with larger models DeepSeek-R1-Distill-Qwen-32B and gpt-oss-120b, we see that the ensemble variant actually outperforms the individual \texttt{SE} module for the Math \& Knowledge datasets, demonstrating that the relative performance of the other modules is stronger in the Math \& Knowledge domain with a stronger base LLM. 

\begin{table*}[!t]

    \centering

    \setlength{\tabcolsep}{1pt}

    \renewcommand{\arraystretch}{0.72}

    \resizebox{0.8\linewidth}{!}{

    \begin{tabular}{lcccc|cccc|cccc|c}

         \toprule[1.5pt]

         \multirow{2}{*}{\textbf{Method}} & \multicolumn{4}{c}{\textbf{Math \& Knowledge}} & \multicolumn{4}{c}{\textbf{Comprehension}} & \multicolumn{4}{c}{\textbf{Biases \& Safety}} & \multirow{2}{*}{\textbf{Overall}} \\ 

          & \small MMLU & \small GSM & \small UMWP & \small \cellcolor{mygray} Overall & \small KC & \small HS & \small Qu & \small \cellcolor{mygray}Overall & \small Mis & \small Prop & \small BBQ & \small \cellcolor{mygray}Overall & \\ \midrule[0.75pt]
         \multicolumn{14}{c}{\textit{\ \ \textbf{\textsc{phi-4}}} }\\ \midrule[0.75pt]
         \textcolor{OliveGreen}{\texttt{SE}} & \textbf{.876} & \textbf{.798} & \textbf{.865} & \cellcolor{mygray} \textbf{.846} & .538 & .727 & .619 & \cellcolor{mygray}.628 & .567 & .608 & \underline{.656} & \cellcolor{mygray} .610 & \cellcolor{myblue}\underline{.695}\\
         \textcolor{OliveGreen}{\texttt{TrInv-LLM}} & .710 & \underline{.743} & .752 & \cellcolor{mygray} \underline{.735} & \textbf{.684} & \underline{.774} & \textbf{.723} & \cellcolor{mygray}\underline{.727} & .555 & .556 & .639 & \cellcolor{mygray} .584 & \cellcolor{myblue}.682\\
         \textcolor{OliveGreen}{\texttt{GROUND}} & .638 & .721 & .556 & \cellcolor{mygray} .638 & .658 & .760 & .609 & \cellcolor{mygray}.675 & \textbf{.696} & \textbf{.826} & \textbf{.772} & \cellcolor{mygray} \textbf{.765} & \cellcolor{myblue}.693\\
         \midrule[0.75pt]
         \textcolor{OliveGreen}{\textsc{Trace Inversion}} & \underline{.712} & .733 & \underline{.757} & \cellcolor{mygray} .734 & \underline{.663} & \textbf{.830} & \underline{.694} & \cellcolor{mygray}\textbf{.729} & \underline{.649} & \underline{.710} & .614 & \cellcolor{mygray} \underline{.658} & \cellcolor{myblue}\textbf{.707}\\
         \midrule[0.75pt]
         \multicolumn{14}{c}{\textit{\ \ \textbf{\textsc{Qwen2.5-32B}}} }\\ \midrule[0.75pt]
         \textcolor{OliveGreen}{\texttt{SE}} & \textbf{.855} & \textbf{.902} & \textbf{.795} & \cellcolor{mygray} \textbf{.851} & .473 & .698 & .638 & \cellcolor{mygray}.603 & .570 & .632 & .579 & \cellcolor{mygray} .594 & \cellcolor{myblue}.683\\
         \textcolor{OliveGreen}{\texttt{TrInv-LLM}} & \underline{.737} & .763 & .707 & \cellcolor{mygray} .736 & \underline{.727} & \textbf{.798} & \textbf{.724} & \cellcolor{mygray}\textbf{.750} & .580 & .623 & .659 & \cellcolor{mygray} .621 & \cellcolor{myblue}\underline{.702}\\
         \textcolor{OliveGreen}{\texttt{GROUND}} & .702 & .775 & .583 & \cellcolor{mygray} .687 & .662 & \underline{.784} & .535 & \cellcolor{mygray}.660 & \textbf{.691} & \textbf{.748} & \textbf{.753} & \cellcolor{mygray} \textbf{.731} & \cellcolor{myblue}.693\\
         \midrule[0.75pt]
         \textcolor{OliveGreen}{\textsc{Trace Inversion}} & .719 & \underline{.850} & \underline{.788} & \cellcolor{mygray} \underline{.786} & \textbf{.789} & .712 & \underline{.717} & \cellcolor{mygray}\underline{.739} & \underline{.670} & \underline{.734} & \underline{.668} & \cellcolor{mygray} \underline{.691} & \cellcolor{myblue}\textbf{.739}\\
         \midrule[0.75pt]
         \multicolumn{14}{c}{\textit{\ \ \textbf{\textsc{DeepSeek-R1-Distill-Qwen-32B}}} }\\ \midrule[0.75pt]
         \textcolor{OliveGreen}{\texttt{SE}} & \underline{.812} & .823 & \underline{.800} & \cellcolor{mygray} \underline{.812} & .552 & .770 & .558 & \cellcolor{mygray}.627 & .545 & .636 & \underline{.627} & \cellcolor{mygray} .603 & \cellcolor{myblue}.680\\
         \textcolor{OliveGreen}{\texttt{TrInv-LLM}} & .771 & \underline{.856} & .720 & \cellcolor{mygray} .782 & \textbf{.724} & \underline{.811} & \textbf{.725} & \cellcolor{mygray}\textbf{.753} & .552 & .580 & .603 & \cellcolor{mygray} .578 & \cellcolor{myblue}\underline{.705}\\
         \textcolor{OliveGreen}{\texttt{GROUND}} & .692 & .764 & .524 & \cellcolor{mygray} .660 & \underline{.657} & \textbf{.816} & .611 & \cellcolor{mygray}\underline{.695} & \underline{.627} & \textbf{.818} & \textbf{.814} & \cellcolor{mygray} \textbf{.753} & \cellcolor{myblue}.703\\
         \midrule[0.75pt]
         \textcolor{OliveGreen}{\textsc{Trace Inversion}} & \textbf{.915} & \textbf{.883} & \textbf{.812} & \cellcolor{mygray} \textbf{.870} & .616 & .731 & \underline{.612} & \cellcolor{mygray}.653 & \textbf{.713} & \underline{.712} & .602 & \cellcolor{mygray} \underline{.676} & \cellcolor{myblue}\textbf{.733}\\
         \midrule[0.75pt]
         \multicolumn{14}{c}{\textit{\ \ \textbf{\textsc{gpt-oss-120b}}} }\\ \midrule[0.75pt]
         \textcolor{OliveGreen}{\texttt{SE}} & \underline{.849} & \underline{.841} & \textbf{.896} & \cellcolor{mygray} \textbf{.862} & .569 & .689 & .605 & \cellcolor{mygray}.621 & .537 & .592 & .601 & \cellcolor{mygray} .577 & \cellcolor{myblue}.687\\
         \textcolor{OliveGreen}{\texttt{TrInv-LLM}} & .778 & .777 & .701 & \cellcolor{mygray} .752 & \underline{.681} & \underline{.769} & \textbf{.657} & \cellcolor{mygray}\underline{.702} & .589 & .632 & .595 & \cellcolor{mygray} .605 & \cellcolor{myblue}.687\\
         \textcolor{OliveGreen}{\texttt{GROUND}} & .704 & .729 & .561 & \cellcolor{mygray} .664 & \textbf{.767} & .761 & .606 & \cellcolor{mygray}\textbf{.711} & \underline{.653} & \textbf{.808} & \textbf{.820} & \cellcolor{mygray} \textbf{.760} & \cellcolor{myblue}\underline{.712}\\
         \midrule[0.75pt]
         \textcolor{OliveGreen}{\textsc{Trace Inversion}} & \textbf{.885} & \textbf{.851} & \underline{.778} & \cellcolor{mygray} \underline{.838} & .585 & \textbf{.814} & \underline{.640} & \cellcolor{mygray}.680 & \textbf{.711} & \underline{.804} & \underline{.695} & \cellcolor{mygray} \underline{.737} & \cellcolor{myblue}\textbf{.751}\\
         \midrule[0.75pt]

         \bottomrule[1.5pt]

    \end{tabular}

    }

    \caption{Ablation of \textsc{Trace Inversion} by looking at the Abstain Accuracy performance of the method with individual distance metrics. Each number reported is the average of three seeds. Performance is shown for each of the four LLMs. Best results in \textbf{bold} and second best in \underline{underline}.}

    \label{tab:ablation_unaggregated}

\end{table*}

\subsection{Complete Baselines with CoT Analysis}
\label{appendix:completeablation2}
\begin{table*}[!h]

    \centering

    \setlength{\tabcolsep}{1pt}

    \renewcommand{\arraystretch}{0.82}

    \resizebox{0.9\textwidth}{!}{

    \begin{tabular}{llcccc|cccc|cccc|c}

         \toprule[1.5pt]

         \multirow{2}{*}{\textbf{Method}} & \multirow{2}{*}{\textbf{Variant}} & \multicolumn{4}{c}{\textbf{Math \& Knowledge}} & \multicolumn{4}{c}{\textbf{Comprehension}} & \multicolumn{4}{c}{\textbf{Biases \& Safety}} & \multirow{2}{*}{\textbf{Overall}} \\ 

          & & \small MMLU & \small GSM & \small UMWP & \small\cellcolor{mygray} Overall & \small KC & \small HS & \small Qu & \small\cellcolor{mygray}Overall & \small Mis & \small Prop & \small BBQ & \small\cellcolor{mygray} Overall & \\ \midrule[0.75pt]
         \multicolumn{15}{c}{\textit{\ \ \textbf{\textsc{phi-4}}} }\\ \midrule[0.75pt]
         \multirow{2}{*}{\textcolor{NavyBlue}{\textsc{Probs}}} & \textit{Regular} & .477 & .509 & .488 & \cellcolor{mygray} .491 & .451 & .666 & .303 & \cellcolor{mygray}.473 & .512 & .624 & .332 & \cellcolor{mygray} .489 & \cellcolor{myblue}.485\\
          & \textit{+ CoT prompt} & \textcolor{myred}{.449} & \textcolor{myred}{.426} & \textcolor{myred}{.395} & \cellcolor{mygray} \textcolor{myred}{.423} & \textcolor{myred}{.424} & \textcolor{myred}{.548} & \textcolor{myred}{.286} & \cellcolor{mygray}\textcolor{myred}{.419} & \textcolor{myred}{.404} & \textcolor{myred}{.559} & \textcolor{myred}{.236} & \cellcolor{mygray} \textcolor{myred}{.400} & \cellcolor{myblue}\textcolor{myred}{.414}\\
         \cmidrule[0.3pt]{1-15}
         \multirow{2}{*}{\textcolor{NavyBlue}{\textsc{AskCali}}} & \textit{Regular} & .471 & .504 & .506 & \cellcolor{mygray} .494 & .550 & .612 & .307 & \cellcolor{mygray}.490 & \textcolor{myred}{.552} & .618 & .299 & \cellcolor{mygray} .490 & \cellcolor{myblue}.491\\
          & \textit{+ CoT prompt} & \textcolor{myred}{.390} & \textcolor{myred}{.394} & \textcolor{myred}{.494} & \cellcolor{mygray} \textcolor{myred}{.426} & \textcolor{myred}{.475} & \textcolor{myred}{.565} & \textcolor{myred}{.169} & \cellcolor{mygray}\textcolor{myred}{.403} & .574 & \textcolor{myred}{.613} & \textcolor{myred}{.220} & \cellcolor{mygray} \textcolor{myred}{.469} & \cellcolor{myblue}\textcolor{myred}{.432}\\
         \cmidrule[0.3pt]{1-15}
         \multirow{2}{*}{\textcolor{Dandelion}{\textsc{Reflect}}} & \textit{Regular} & .379 & .541 & .438 & \cellcolor{mygray} .453 & .633 & \textcolor{myred}{.771} & .405 & \cellcolor{mygray}.603 & \textcolor{myred}{.455} & .515 & .411 & \cellcolor{mygray} .460 & \cellcolor{myblue}.505\\
          & \textit{+ CoT prompt} & \textcolor{myred}{.278} & \textcolor{myred}{.487} & \textcolor{myred}{.380} & \cellcolor{mygray} \textcolor{myred}{.381} & \textcolor{myred}{.556} & .794 & \textcolor{myred}{.388} & \cellcolor{mygray}\textcolor{myred}{.579} & .495 & \textcolor{myred}{.411} & \textcolor{myred}{.271} & \cellcolor{mygray} \textcolor{myred}{.392} & \cellcolor{myblue}\textcolor{myred}{.451}\\
         \cmidrule[0.3pt]{1-15}
         \multirow{2}{*}{\textcolor{Maroon}{\textsc{Cooperate}}} & \textit{Regular} & \textcolor{myred}{.424} & .685 & .420 & \cellcolor{mygray} .510 & .504 & .718 & .414 & \cellcolor{mygray}.545 & \textcolor{myred}{.369} & \textcolor{myred}{.598} & .422 & \cellcolor{mygray} \textcolor{myred}{.463} & \cellcolor{myblue}.506\\
          & \textit{+ CoT prompt} & .442 & \textcolor{myred}{.643} & \textcolor{myred}{.385} & \cellcolor{mygray} \textcolor{myred}{.490} & \textcolor{myred}{.408} & \textcolor{myred}{.691} & \textcolor{myred}{.394} & \cellcolor{mygray}\textcolor{myred}{.498} & .400 & .601 & \textcolor{myred}{.398} & \cellcolor{mygray} .466 & \cellcolor{myblue}\textcolor{myred}{.485}\\
         \cmidrule[0.3pt]{1-15}
         \multirow{2}{*}{\textcolor{Maroon}{\textsc{Compete}}} & \textit{Regular} & .578 & .547 & \textcolor{myred}{.516} & \cellcolor{mygray} .547 & .426 & .690 & .533 & \cellcolor{mygray}.550 & .467 & .600 & .312 & \cellcolor{mygray} .460 & \cellcolor{myblue}.519\\
          & \textit{+ CoT prompt} & \textcolor{myred}{.507} & \textcolor{myred}{.424} & .558 & \cellcolor{mygray} \textcolor{myred}{.497} & \textcolor{myred}{.403} & \textcolor{myred}{.669} & \textcolor{myred}{.468} & \cellcolor{mygray}\textcolor{myred}{.513} & \textcolor{myred}{.423} & \textcolor{myred}{.508} & \textcolor{myred}{.279} & \cellcolor{mygray} \textcolor{myred}{.404} & \cellcolor{myblue}\textcolor{myred}{.471}\\
         \midrule[0.75pt]
         \multicolumn{15}{c}{\textit{\ \ \textbf{\textsc{Qwen2.5-32B}}} }\\ \midrule[0.75pt]
         \multirow{2}{*}{\textcolor{NavyBlue}{\textsc{Probs}}} & \textit{Regular} & .741 & .711 & .512 & \cellcolor{mygray} .655 & \textcolor{myred}{.329} & .551 & .397 & \cellcolor{mygray}.426 & .456 & .641 & .498 & \cellcolor{mygray} .532 & \cellcolor{myblue}.537\\
          & \textit{+ CoT prompt} & \textcolor{myred}{.703} & \textcolor{myred}{.577} & \textcolor{myred}{.418} & \cellcolor{mygray} \textcolor{myred}{.566} & .339 & \textcolor{myred}{.501} & \textcolor{myred}{.381} & \cellcolor{mygray}\textcolor{myred}{.407} & \textcolor{myred}{.394} & \textcolor{myred}{.581} & \textcolor{myred}{.416} & \cellcolor{mygray} \textcolor{myred}{.464} & \cellcolor{myblue}\textcolor{myred}{.479}\\
         \cmidrule[0.3pt]{1-15}
         \multirow{2}{*}{\textcolor{NavyBlue}{\textsc{AskCali}}} & \textit{Regular} & .711 & .684 & \textcolor{myred}{.601} & \cellcolor{mygray} .665 & .473 & .513 & .319 & \cellcolor{mygray}.435 & .451 & .647 & .401 & \cellcolor{mygray} .500 & \cellcolor{myblue}.533\\
          & \textit{+ CoT prompt} & \textcolor{myred}{.682} & \textcolor{myred}{.582} & .619 & \cellcolor{mygray} \textcolor{myred}{.628} & \textcolor{myred}{.426} & \textcolor{myred}{.463} & \textcolor{myred}{.230} & \cellcolor{mygray}\textcolor{myred}{.373} & \textcolor{myred}{.443} & \textcolor{myred}{.625} & \textcolor{myred}{.323} & \cellcolor{mygray} \textcolor{myred}{.464} & \cellcolor{myblue}\textcolor{myred}{.488}\\
         \cmidrule[0.3pt]{1-15}
         \multirow{2}{*}{\textcolor{Dandelion}{\textsc{Reflect}}} & \textit{Regular} & .689 & .731 & .639 & \cellcolor{mygray} .686 & .618 & \textcolor{myred}{.610} & .305 & \cellcolor{mygray}.511 & .415 & .621 & .411 & \cellcolor{mygray} .482 & \cellcolor{myblue}.560\\
          & \textit{+ CoT prompt} & \textcolor{myred}{.646} & \textcolor{myred}{.669} & \textcolor{myred}{.549} & \cellcolor{mygray} \textcolor{myred}{.622} & \textcolor{myred}{.536} & .625 & \textcolor{myred}{.236} & \cellcolor{mygray}\textcolor{myred}{.466} & \textcolor{myred}{.382} & \textcolor{myred}{.562} & \textcolor{myred}{.334} & \cellcolor{mygray} \textcolor{myred}{.426} & \cellcolor{myblue}\textcolor{myred}{.504}\\
         \cmidrule[0.3pt]{1-15}
         \multirow{2}{*}{\textcolor{Maroon}{\textsc{Cooperate}}} & \textit{Regular} & \textcolor{myred}{.637} & .725 & .420 & \cellcolor{mygray} .594 & .516 & .618 & .345 & \cellcolor{mygray}.493 & \textcolor{myred}{.322} & .603 & .424 & \cellcolor{mygray} .450 & \cellcolor{myblue}.512\\
          & \textit{+ CoT prompt} & .639 & \textcolor{myred}{.674} & \textcolor{myred}{.406} & \cellcolor{mygray} \textcolor{myred}{.573} & \textcolor{myred}{.481} & \textcolor{myred}{.608} & \textcolor{myred}{.288} & \cellcolor{mygray}\textcolor{myred}{.459} & .348 & \textcolor{myred}{.544} & \textcolor{myred}{.414} & \cellcolor{mygray} \textcolor{myred}{.436} & \cellcolor{myblue}\textcolor{myred}{.489}\\
         \cmidrule[0.3pt]{1-15}
         \multirow{2}{*}{\textcolor{Maroon}{\textsc{Compete}}} & \textit{Regular} & .688 & .747 & .656 & \cellcolor{mygray} .697 & \textcolor{myred}{.509} & .672 & \textcolor{myred}{.488} & \cellcolor{mygray}.556 & .667 & .704 & .490 & \cellcolor{mygray} .620 & \cellcolor{myblue}.625\\
          & \textit{+ CoT prompt} & \textcolor{myred}{.629} & \textcolor{myred}{.590} & \textcolor{myred}{.623} & \cellcolor{mygray} \textcolor{myred}{.614} & .545 & \textcolor{myred}{.614} & .492 & \cellcolor{mygray}\textcolor{myred}{.551} & \textcolor{myred}{.596} & \textcolor{myred}{.608} & \textcolor{myred}{.427} & \cellcolor{mygray} \textcolor{myred}{.544} & \cellcolor{myblue}\textcolor{myred}{.569}\\
         \midrule[0.75pt]
         \multicolumn{15}{c}{\textit{\ \ \textbf{\textsc{DeepSeek-R1-Distill-Qwen-32B}}} }\\ \midrule[0.75pt]
         \multirow{2}{*}{\textcolor{NavyBlue}{\textsc{Probs}}} & \textit{Regular} & .770 & .739 & .600 & \cellcolor{mygray} .703 & .653 & .622 & .412 & \cellcolor{mygray}.562 & .503 & .644 & .487 & \cellcolor{mygray} .545 & \cellcolor{myblue}.603\\
          & \textit{+ CoT prompt} & \textcolor{myred}{.706} & \textcolor{myred}{.594} & \textcolor{myred}{.528} & \cellcolor{mygray} \textcolor{myred}{.610} & \textcolor{myred}{.622} & \textcolor{myred}{.501} & \textcolor{myred}{.404} & \cellcolor{mygray}\textcolor{myred}{.509} & \textcolor{myred}{.368} & \textcolor{myred}{.618} & \textcolor{myred}{.411} & \cellcolor{mygray} \textcolor{myred}{.466} & \cellcolor{myblue}\textcolor{myred}{.528}\\
         \cmidrule[0.3pt]{1-15}
         \multirow{2}{*}{\textcolor{NavyBlue}{\textsc{AskCali}}} & \textit{Regular} & .765 & .784 & .601 & \cellcolor{mygray} .717 & .557 & .454 & .317 & \cellcolor{mygray}.443 & .511 & .672 & .404 & \cellcolor{mygray} .529 & \cellcolor{myblue}.563\\
          & \textit{+ CoT prompt} & \textcolor{myred}{.735} & \textcolor{myred}{.687} & \textcolor{myred}{.568} & \cellcolor{mygray} \textcolor{myred}{.663} & \textcolor{myred}{.498} & \textcolor{myred}{.445} & \textcolor{myred}{.198} & \cellcolor{mygray}\textcolor{myred}{.380} & \textcolor{myred}{.506} & \textcolor{myred}{.658} & \textcolor{myred}{.335} & \cellcolor{mygray} \textcolor{myred}{.500} & \cellcolor{myblue}\textcolor{myred}{.514}\\
         \cmidrule[0.3pt]{1-15}
         \multirow{2}{*}{\textcolor{Dandelion}{\textsc{Reflect}}} & \textit{Regular} & .748 & .744 & .639 & \cellcolor{mygray} .710 & \textcolor{myred}{.633} & .611 & .510 & \cellcolor{mygray}.585 & \textcolor{myred}{.510} & .618 & .310 & \cellcolor{mygray} .479 & \cellcolor{myblue}.591\\
          & \textit{+ CoT prompt} & \textcolor{myred}{.679} & \textcolor{myred}{.637} & \textcolor{myred}{.537} & \cellcolor{mygray} \textcolor{myred}{.618} & .637 & \textcolor{myred}{.582} & \textcolor{myred}{.429} & \cellcolor{mygray}\textcolor{myred}{.550} & .519 & \textcolor{myred}{.496} & \textcolor{myred}{.192} & \cellcolor{mygray} \textcolor{myred}{.402} & \cellcolor{myblue}\textcolor{myred}{.523}\\
         \cmidrule[0.3pt]{1-15}
         \multirow{2}{*}{\textcolor{Maroon}{\textsc{Cooperate}}} & \textit{Regular} & \textcolor{myred}{.849} & .715 & \textcolor{myred}{.722} & \cellcolor{mygray} .762 & .707 & .718 & .298 & \cellcolor{mygray}.574 & .488 & .511 & .420 & \cellcolor{mygray} .473 & \cellcolor{myblue}.603\\
          & \textit{+ CoT prompt} & .861 & \textcolor{myred}{.593} & .755 & \cellcolor{mygray} \textcolor{myred}{.736} & \textcolor{myred}{.594} & \textcolor{myred}{.690} & \textcolor{myred}{.241} & \cellcolor{mygray}\textcolor{myred}{.508} & \textcolor{myred}{.484} & \textcolor{myred}{.444} & \textcolor{myred}{.396} & \cellcolor{mygray} \textcolor{myred}{.441} & \cellcolor{myblue}\textcolor{myred}{.562}\\
         \cmidrule[0.3pt]{1-15}
         \multirow{2}{*}{\textcolor{Maroon}{\textsc{Compete}}} & \textit{Regular} & .784 & .647 & .556 & \cellcolor{mygray} .662 & .329 & .488 & .301 & \cellcolor{mygray}.373 & .583 & .605 & .338 & \cellcolor{mygray} .509 & \cellcolor{myblue}.515\\
          & \textit{+ CoT prompt} & \textcolor{myred}{.731} & \textcolor{myred}{.570} & \textcolor{myred}{.514} & \cellcolor{mygray} \textcolor{myred}{.605} & \textcolor{myred}{.313} & \textcolor{myred}{.466} & \textcolor{myred}{.297} & \cellcolor{mygray}\textcolor{myred}{.359} & \textcolor{myred}{.465} & \textcolor{myred}{.547} & \textcolor{myred}{.317} & \cellcolor{mygray} \textcolor{myred}{.443} & \cellcolor{myblue}\textcolor{myred}{.469}\\
         \midrule[0.75pt]
         \multicolumn{15}{c}{\textit{\ \ \textbf{\textsc{gpt-oss-120b}}} }\\ \midrule[0.75pt]
         \multirow{2}{*}{\textcolor{NavyBlue}{\textsc{Probs}}} & \textit{Regular} & .767 & .799 & .577 & \cellcolor{mygray} .714 & .617 & .632 & .417 & \cellcolor{mygray}.555 & .614 & .624 & .408 & \cellcolor{mygray} .549 & \cellcolor{myblue}.606\\
          & \textit{+ CoT prompt} & \textcolor{myred}{.738} & \textcolor{myred}{.731} & \textcolor{myred}{.511} & \cellcolor{mygray} \textcolor{myred}{.660} & \textcolor{myred}{.571} & \textcolor{myred}{.522} & \textcolor{myred}{.417} & \cellcolor{mygray}\textcolor{myred}{.503} & \textcolor{myred}{.550} & \textcolor{myred}{.526} & \textcolor{myred}{.284} & \cellcolor{mygray} \textcolor{myred}{.453} & \cellcolor{myblue}\textcolor{myred}{.539}\\
         \cmidrule[0.3pt]{1-15}
         \multirow{2}{*}{\textcolor{NavyBlue}{\textsc{AskCali}}} & \textit{Regular} & .725 & .804 & \textcolor{myred}{.501} & \cellcolor{mygray} .677 & .552 & .682 & .414 & \cellcolor{mygray}.549 & \textcolor{myred}{.588} & \textcolor{myred}{.590} & .426 & \cellcolor{mygray} .535 & \cellcolor{myblue}.587\\
          & \textit{+ CoT prompt} & \textcolor{myred}{.653} & \textcolor{myred}{.741} & .532 & \cellcolor{mygray} \textcolor{myred}{.642} & \textcolor{myred}{.497} & \textcolor{myred}{.647} & \textcolor{myred}{.344} & \cellcolor{mygray}\textcolor{myred}{.496} & .604 & .592 & \textcolor{myred}{.343} & \cellcolor{mygray} \textcolor{myred}{.513} & \cellcolor{myblue}\textcolor{myred}{.550}\\
         \cmidrule[0.3pt]{1-15}
         \multirow{2}{*}{\textcolor{Dandelion}{\textsc{Reflect}}} & \textit{Regular} & .759 & .739 & .533 & \cellcolor{mygray} .677 & .438 & .623 & .305 & \cellcolor{mygray}.455 & \textcolor{myred}{.615} & .644 & .413 & \cellcolor{mygray} .557 & \cellcolor{myblue}.563\\
          & \textit{+ CoT prompt} & \textcolor{myred}{.641} & \textcolor{myred}{.718} & \textcolor{myred}{.462} & \cellcolor{mygray} \textcolor{myred}{.607} & \textcolor{myred}{.355} & \textcolor{myred}{.602} & \textcolor{myred}{.227} & \cellcolor{mygray}\textcolor{myred}{.394} & .632 & \textcolor{myred}{.559} & \textcolor{myred}{.335} & \cellcolor{mygray} \textcolor{myred}{.509} & \cellcolor{myblue}\textcolor{myred}{.503}\\
         \cmidrule[0.3pt]{1-15}
         \multirow{2}{*}{\textcolor{Maroon}{\textsc{Cooperate}}} & \textit{Regular} & \textcolor{myred}{.749} & .817 & \textcolor{myred}{.520} & \cellcolor{mygray} .695 & .704 & .812 & .312 & \cellcolor{mygray}.609 & .607 & \textcolor{myred}{.590} & .388 & \cellcolor{mygray} .528 & \cellcolor{myblue}.611\\
          & \textit{+ CoT prompt} & .754 & \textcolor{myred}{.754} & .534 & \cellcolor{mygray} \textcolor{myred}{.681} & \textcolor{myred}{.585} & \textcolor{myred}{.767} & \textcolor{myred}{.266} & \cellcolor{mygray}\textcolor{myred}{.539} & \textcolor{myred}{.584} & .604 & \textcolor{myred}{.299} & \cellcolor{mygray} \textcolor{myred}{.496} & \cellcolor{myblue}\textcolor{myred}{.572}\\
         \cmidrule[0.3pt]{1-15}
         \multirow{2}{*}{\textcolor{Maroon}{\textsc{Compete}}} & \textit{Regular} & .788 & .847 & \textcolor{myred}{.614} & \cellcolor{mygray} .750 & .616 & .691 & \textcolor{myred}{.590} & \cellcolor{mygray}.632 & .588 & .611 & .487 & \cellcolor{mygray} .562 & \cellcolor{myblue}.648\\
          & \textit{+ CoT prompt} & \textcolor{myred}{.705} & \textcolor{myred}{.763} & .636 & \cellcolor{mygray} \textcolor{myred}{.701} & \textcolor{myred}{.575} & \textcolor{myred}{.655} & .590 & \cellcolor{mygray}\textcolor{myred}{.607} & \textcolor{myred}{.456} & \textcolor{myred}{.533} & \textcolor{myred}{.413} & \cellcolor{mygray} \textcolor{myred}{.467} & \cellcolor{myblue}\textcolor{myred}{.592}\\
         \midrule[0.75pt]

         \bottomrule[1.5pt]

    \end{tabular}

    }

    \caption{Abstain Accuracy of abstain strategies on nine datasets for each of the four LLMs. Each number reported is the average of three seeds. We group the nine datasets (MMLU, GSM, UMWP, K-Crosswords, Hellaswag, Quail, Misconceptions, Propaganda, and BBQ) according to domain and provide aggregate results per domain and across all datasets. Approaches are color-coded per category: \textcolor{NavyBlue}{calibration}, \textcolor{Dandelion}{prompting}, \textcolor{Maroon}{collaboration}, and \textcolor{OliveGreen}{our method}.}

    \label{tab:big_unaggregated_averaged}

    \vspace*{-10pt}

\end{table*}
Lastly, we provide a complete table showing the degradation of Abstain Accuracy when using CoT prompts with the baselines (see Table \ref{tab:big_unaggregated_averaged}). The disaggregated results highlight the extent of CoT degradation in certain settings. For example, the  \textsc{Reflect} method demonstrates a  11.8\% decrease in Abstain Accuracy for gpt-oss-120b on the MMLU dataset and 14.0\% decrease for phi-4 on the BBQ dataset.  

\end{document}